%% file: iclr2027_conference.tex
\documentclass{article}
\usepackage{iclr2027_conference,times}

\input{math_commands.tex}

\usepackage{fontawesome5}
\usepackage{hyperref}
\usepackage{url}
\usepackage{xcolor}
\usepackage{booktabs}
\usepackage{longtable}
\usepackage{pdflscape}
\usepackage{array}
\usepackage{amsmath}
\usepackage{enumitem}
\usepackage{graphicx}
\usepackage{soul}
\usepackage{ragged2e}
\usepackage{tabularx,multirow}
\usepackage{makecell}
\usepackage{wrapfig}

\definecolor{qScenario}{RGB}{218,235,252}
\definecolor{qMechanism}{RGB}{220,241,225}
\definecolor{qField}{RGB}{255,236,204}
\definecolor{qConfig}{RGB}{237,226,247}
\definecolor{qObservable}{RGB}{252,222,226}
\definecolor{lightgrayrule}{RGB}{205,205,205}

\definecolor{revcolor}{HTML}{C1440E}

\definecolor{revcolorb}{HTML}{0072B2}

\newcommand{\slot}[1]{\underline{#1}}
\soulregister\slot7

\newcommand{\qscenario}[1]{{\sethlcolor{qScenario}\hl{#1}}}
\newcommand{\qmechanism}[1]{{\sethlcolor{qMechanism}\hl{#1}}}
\newcommand{\qfield}[1]{{\sethlcolor{qField}\hl{#1}}}
\newcommand{\qconfig}[1]{{\sethlcolor{qConfig}\hl{#1}}}
\newcommand{\qobs}[1]{{\sethlcolor{qObservable}\hl{#1}}}

\title{World Embedding Benchmark}

\vspace{10pt}

\author{%
{\fontsize{10.5pt}{11.5pt}\selectfont
\begin{tabular}{@{}l@{}}
\textbf{Yiqi Liu},
\textbf{Ruifeng Yuan},
\textbf{Yang Wang},
\textbf{Long Li},
\textbf{Fengyu Cai},
\textbf{Hou Pong Chan},
\textbf{Jialin Yu},\\
\textbf{Hao Zhang},
\textbf{Chenghua Lin},
\textbf{Chenghao Xiao\textsuperscript{\textdagger}}
\end{tabular}
}
\\[10pt]
{\fontsize{9.5pt}{11pt}\selectfont
World-Embedding Team
}
}

\iclrfinalcopy
\begin{document}

\maketitle

\begingroup
\renewcommand{\thefootnote}{\fnsymbol{footnote}}

\footnotetext[2]{Corresponding author: \href{mailto:xiaochenghao@sufe.edu.cn}{xiaochenghao@sufe.edu.cn}.}

\endgroup

\vspace{-25pt}
\begin{center}
\footnotesize
\begin{tabular}{@{}cl@{}}
\raisebox{-0.15em}{\includegraphics[height=1.05em]{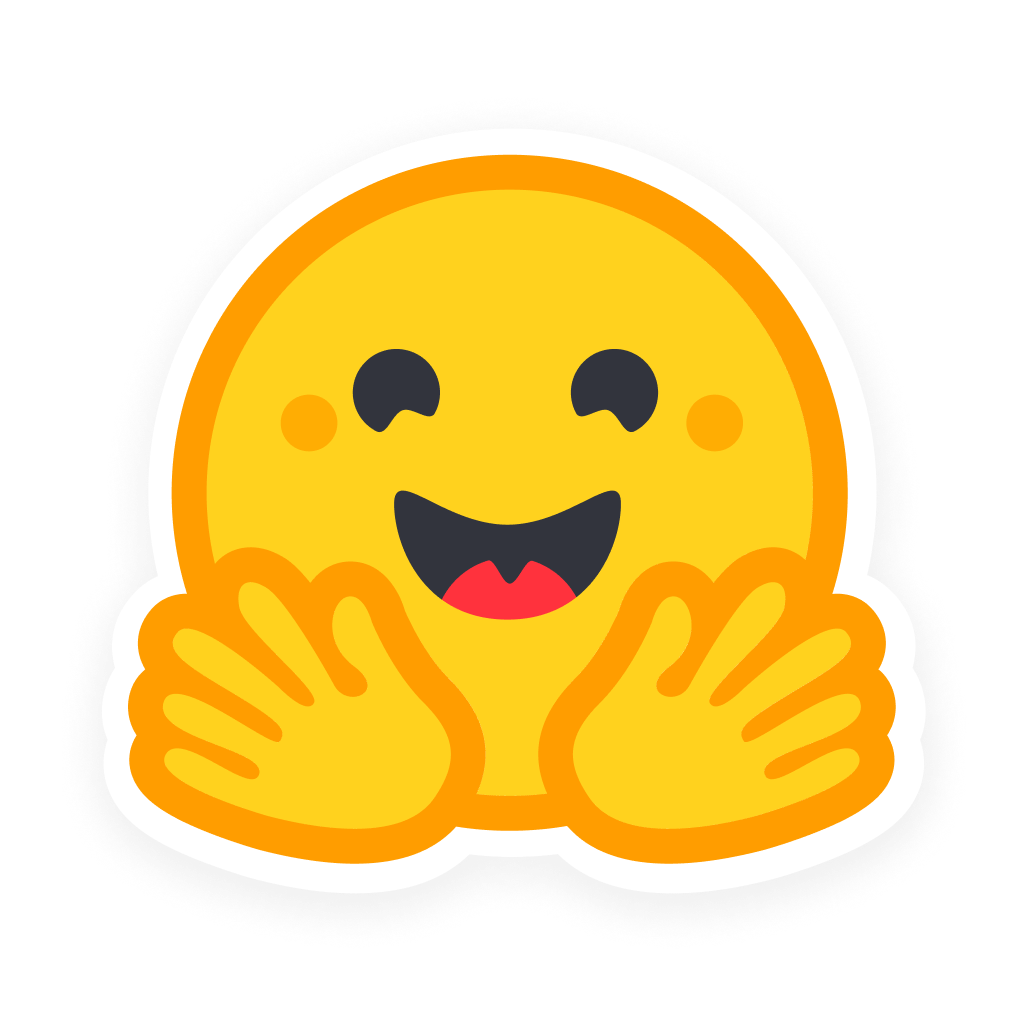}}
&
\href{https://huggingface.co/collections/World-Representation-Lab/world-embedding-benchmark}
{\texttt{huggingface.co/World-Representation-Lab/world-embedding-benchmark}}
\\[1pt]
\raisebox{-0.08em}{\faGithub}
&
\href{https://github.com/World-Representation-Lab/World-Embedding-Benchmark}
{\texttt{github.com/World-Representation-Lab/World-Embedding-Benchmark}}
\end{tabular}
\end{center}
\vspace{10pt}

\begin{abstract}
Physical fidelity has received increasing attention in world models and video generation, yet how video representations encode physical information remains less understood. We introduce the World Embedding Benchmark, comprising 8,000 controlled simulation cases from 80 families spanning fluid mechanics, solid mechanics, dynamics, and optics \& electromagnetism. Each case pairs a rendered video with simulation-derived physical annotations, supporting three complementary tasks: text-video retrieval, physical-property regression, and multiple-choice video-description pair classification. We use these tasks to distinguish cross-modal physical alignment from the recoverability of quantitative physical information. Evaluated pre-trained omnimodal embedding models show weak retrieval and near-chance within-family pair classification, while lightweight probes recover useful physical information from frozen video embeddings. Continual contrastive training with physics-specific video-text pairs improves retrieval and pair classification but degrades physical-property regression, revealing a trade-off between alignment and quantitative information recoverability. Finally, we use the embeddings to retrieve reference videos for retrieval-augmented generation with MiniMax-H3. Retrieved references improve the physical fidelity of generated videos, with stronger retrieval models yielding larger gains in our experiments. Together, these findings highlight the need to evaluate physical alignment and property recoverability jointly, and demonstrate the utility of physical representations for improving video generation.
\end{abstract}

\section{Introduction}
Physical fidelity is essential to the ambition of using video generative models as world models: a useful model of the physical world must capture how systems behave under different conditions, beyond producing visually plausible scenes \citep{bansal2025videophy,meng2025towards,kang2025how,li2026worldmodelbench}. Recent evaluations have made this challenge explicit, revealing substantial gaps between visual realism and adherence to physical commonsense in generated videos \citep{bansal2025videophy,meng2025towards,zheng2025vbench2}. This growing attention to physical fidelity raises a critical question about representation learning: \emph{what information about the physical world is preserved in video embeddings, and how accessible is that information?} A representation may identify a scene as fluid flow or elastic deformation yet fail to distinguish changes in viscosity or stiffness. Such distinctions matter when embeddings are used to retrieve physical examples, estimate system properties, or provide reference evidence for generation \citep{peruzzo2025ragme,cheng2026physrag}.

Evaluating physical representations requires separating two capabilities: encoding information about physical quantities and aligning that information with language. A physical property may be recoverable by a trained probe even when embedding similarity cannot match a video to a description of that property. Conversely, successful matching of broad scene categories need not imply sensitivity to the parameters governing their behavior. Existing scientific simulation benchmarks provide rich resources for learning physical dynamics \citep{takamoto2022pdebench,luo2024cfdbench,ohana2024thewell}, but evaluating multimodal embeddings calls for paired video-language observations and quantitative physical targets within a shared, controlled setting. This motivates a benchmark that measures both cross-modal alignment and the recoverability of physical information in video representations.

We introduce the \textbf{World Embedding Benchmark}, a comprehensive benchmark for evaluating physical information in video embeddings. It comprises 8,000 simulation cases from 80 families spanning fluid mechanics, solid mechanics, dynamics, and optics \& electromagnetism. Each family contains 100 instances of a physical system with different parameter settings. We combine branch-specific simulation engines
with analytical formulations and numerical integration where appropriate. These controlled configurations provide rendered videos together with their generating parameters and computed physical quantities, enabling evaluation beyond coarse scene semantics.

The benchmark supports three complementary tasks. \textbf{Text-video retrieval} measures cross-modal matching between physical descriptions and simulation videos.
\textbf{Physical-property regression} measures whether lightweight probes can recover continuous physical quantities from frozen video embeddings. \textbf{Pair classification} asks models to select the correct video-description pair from a small set of candidates. 
Together, these tasks assess cross-modal physical alignment, quantitative information recoverability, and fine-grained physical discrimination.
Using this benchmark, we investigate three research questions.

\paragraph{RQ1: Do current representation models encode physics in video representations?}
We evaluate pre-trained omnimodal embedding models and find a marked discrepancy across tasks: text-video retrieval and within-family pair classification remain close to their respective chance baselines, whereas lightweight regression probes recover useful physical information from frozen video representations. These results suggest that the evaluated embeddings retain signals predictive of physical properties, but that these signals are not readily accessible through cross-modal similarity. Directly prompted MLLMs perform substantially better on pair classification, suggesting that relevant physical capabilities exist but are not fully exposed in embedding space.

\paragraph{RQ2: Can contrastive adaptation improve physical alignment?}
We investigate whether adding physics-specific video-text pairs to continual contrastive training improves the physical alignment of pre-trained embeddings. Adaptation improves retrieval and pair classification, but reduces physical-property regression. This trade-off suggests that organizing embeddings for cross-modal matching can compromise the recoverability of quantitative physical information. Overall, these results suggest that contrastive training primarily reorganizes and aligns existing physical signals in the representation space, rather than substantially enhancing the underlying physical capabilities.

\paragraph{RQ3: Can physical representations improve video generation?}
We use omnimodal embedding models as retrievers to provide physically relevant reference videos for retrieval-augmented generation with MiniMax-H3~\citep{minimaxh32026}. Retrieved references improve the physical fidelity of generated videos, and embeddings with stronger physics retrieval performance yield larger improvements in our experiments. 
The results suggest that improving physical representations offers a practical route to supporting more faithful generation of physical scenes.

Our contributions are three-fold:
(1) We introduce the World Embedding Benchmark, comprising 8,000 controlled simulation cases across 80 physical families and three complementary tasks for evaluating physical representations. 
(2) We identify a gap between physical information encoded by multimodal models and its cross-modal alignment in embedding space, and show that contrastive adaptation improves alignment while degrading physical-property regression.
(3) We demonstrate the downstream utility of physical embeddings for retrieval-augmented video generation, connecting stronger physical retrieval performance to improved physical fidelity of generated videos. More broadly, these results point to physically grounded representations as a useful bridge between understanding physical systems and improving generative world models.

\begin{figure*}[ht!]
    \centering
    \includegraphics[width=\textwidth]{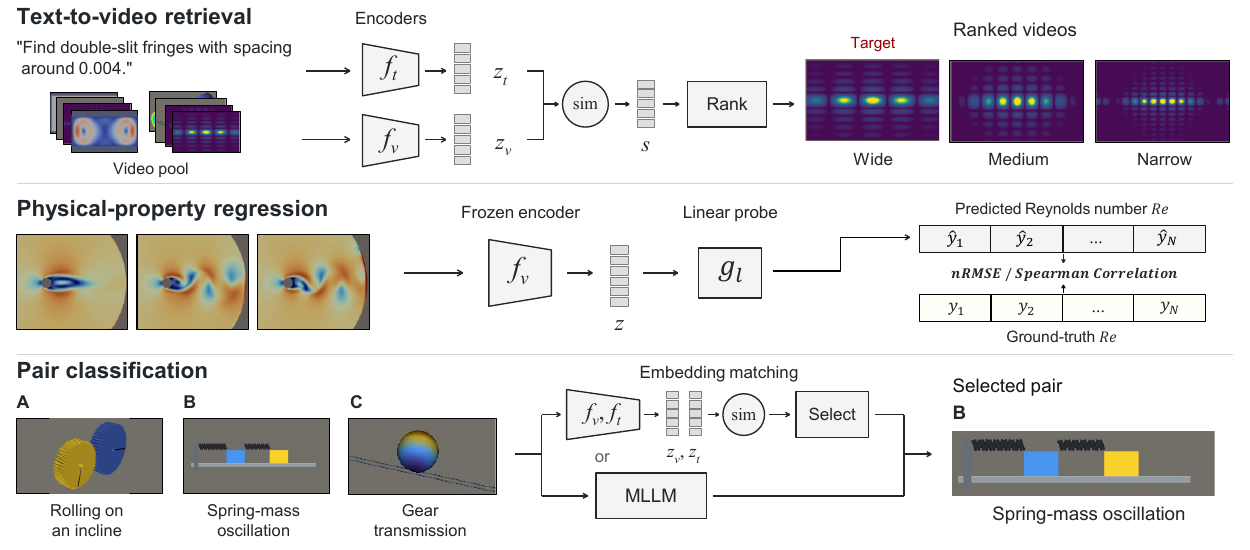}
    \caption{Overview of the three evaluation tasks: text-video retrieval, physical-property regression, and video-description pair classification.}
    \label{fig:task_overview_flow}
\end{figure*}

\section{World Embedding Benchmark}
\label{world_embedding_benchmark}

The World Embedding Benchmark evaluates the extent to which video representations preserve information about physical systems and their underlying parameters. The benchmark is built from 8000 physics simulation cases from 80 families across multiple branches: Fluid Mechanics, Solid Mechanics, Dynamics, and Optics \& Electromagnetism. Each family contains 100 cases, representing the same physical system under different parameter settings.

The simulation cases support three complementary evaluation tasks: 1) \textbf{Text-video retrieval}, which evaluates the alignment between natural-language descriptions of physical systems and their corresponding video representations. 2) \textbf{Physical property regression}, which evaluates whether continuous quantities can be recovered from frozen video representations using lightweight regression models. 3) \textbf{Pair classification}, which evaluates whether models can distinguish the correct video-description correspondence from alternative candidates.
Figure~\ref{fig:task_overview_flow} summarizes the three evaluation tasks, which respectively assess cross-modal physical alignment, quantitative physical information recoverability, and fine-grained physical discrimination.

\input{Tables/physics_dataset_overview}

\subsection{Physics Simulation}
\label{physics_simulation}

\begin{figure*}[t]
    \centering
    \includegraphics[width=\textwidth]{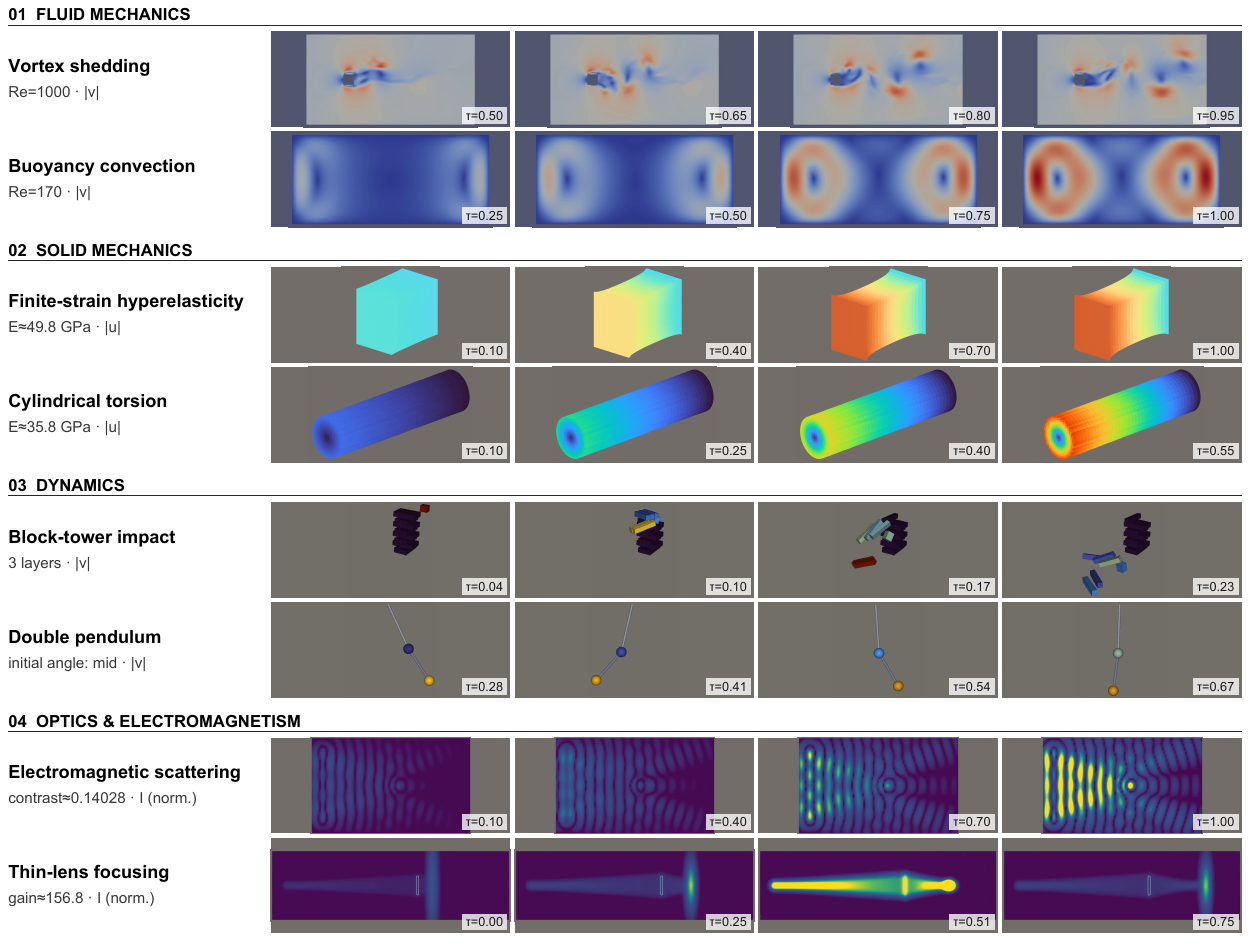}
    \caption{Example video frames of representative scene families of the 4 physics branches, illustrating different stages of each simulation sequence. The full dataset contains videos from 80 families.}
    \label{fig:simulation_dataset_overview}
\end{figure*}

We construct the physics simulation subset from parameterized, representative scene families across Fluid Mechanics, Solid Mechanics, Dynamics, and Optics \& Electromagnetism. Each instance is defined by its governing model, constitutive law, geometry, material parameters, boundary conditions, and initial conditions. Physical responses are generated using OpenFOAM for fluid mechanics \citep{openfoam_software,weller1998tensorial}, FEniCSx/DOLFINx for solid mechanics \citep{baratta2023dolfinx}, Project Chrono for rigid-body and multibody dynamics \citep{tasora2016chrono}, and HCIPy together with Mitsuba~3 for optics and MEEP for electromagnetics \citep{por2018hcipy,Mitsuba3,oskooi2010meep}. The exported fields and trajectories are rendered into PNG frames with ParaView~6.0.1 \citep{ahrens2005paraview} and encoded into MP4 clips with FFmpeg. The rendered clips therefore depict physically faithful responses with quantitative meaning, and generating evaluation data from validated simulators follows a growing line of scientific machine learning benchmarks \citep{takamoto2022pdebench,ohana2024thewell,luo2024cfdbench}. Simulation provides explicit access to the generating parameters and computed physical states, enabling controlled evaluation of physical information in video representations. Representative examples from the four physics branches are shown in Figures~\ref{fig:simulation_dataset_overview}. Detailed formulations, computation methods, case parameters, and field outputs are provided in Appendix~\ref{app:A}.

\paragraph{Quality control and self-checks}
To improve data quality and reduce duplication, we apply automated checks during dataset construction. These checks flag failed or non-convergent runs, NaN or infinite values, and unexpected physical responses. Video checks examine file properties and use sampled-frame statistics to identify blank, nearly static, or flickering videos. We also check case IDs and parameter settings for duplicates.

\subsubsection{Fluid Mechanics}
\label{fluid_mechanics}

The fluid dynamics subset contains 700 videos across seven families instantiated from 15 OpenFOAM configurations \citep{openfoam_software,weller1998tensorial}. The configurations cover internal and external flows, cavity and buoyancy-driven flows, jets, non-Newtonian flow, and passive-scalar transport, including lid-driven cavity and backward-facing step flow \citep{ghia1982cavity,armaly1983bfs}. The case parameters include Reynolds and P\'eclet numbers, velocity, viscosity, geometry, scalar diffusivity, buoyancy parameters, and rheological parameters. Outputs include velocity, temperature, and transported-scalar fields. Further details are provided in Appendix~\ref{app:A2-fluid}.

\subsubsection{Solid Mechanics}

The solid mechanics subset contains 2,700 videos across 27 families generated with FEniCSx/DOLFINx \citep{baratta2023dolfinx}. The families cover linear elasticity, finite-strain hyperelasticity, geometrically nonlinear deformation, indentation, stress concentration, and history-dependent response under tension, compression, bending, and torsion. The cases vary in material properties, geometry, and loading conditions, and the released videos primarily visualize displacement magnitude on the deformed geometry. Detailed formulations are provided in Appendix~\ref{app:A2-solid}.

\subsubsection{Dynamics}

The dynamics subset contains 1,900 videos across 19 families combining Project Chrono \citep{tasora2016chrono} with Analytical / ODE integration. The families cover rigid-body motion, collision and contact, articulated mechanisms, vibration, wave propagation, rolling and sliding, rotating machinery, and structural response. The case parameters include geometry, mass, gravity, initial conditions, restitution and friction coefficients, stiffness and damping, and excitation conditions. Outputs include trajectories and family-specific quantities such as velocity, displacement, angular speed, stress, equivalent plastic strain, and damage. Detailed formulations are provided in Appendix~\ref{app:A2-dynamics}.

\subsubsection{Optics \& Electromagnetism}

The optics and electromagnetism subset contains 2,700 videos across 27 families. Nineteen optics families cover reflection and refraction, diffraction, interference, focusing, Gaussian-beam propagation, dispersion, and polarization, using analytical relations or HCIPy \citep{por2018hcipy}, with Mitsuba~3 \citep{Mitsuba3} used as an auxiliary backend for selected geometric-optics configurations. The remaining eight families use Meep FDTD \citep{oskooi2010meep} for electromagnetic scattering, waveguides, photonic-crystal structures, gratings, dipole radiation, evanescent fields, and dispersive plasmonic systems. Detailed formulations and field definitions are provided in Appendix~\ref{app:A2-optics}.

\subsection{Benchmark Formulation}
\label{benchmark formulation}

\paragraph{Retrieval}

We formulate physics-aware retrieval as bidirectional text-video retrieval within each physics branch. We evaluate both text-to-video and video-to-text retrieval over the full branch corpus and report their mean Recall@K. For example, the Solid Mechanics corpus contains 2,700 paired videos and queries from 27 families.
For each family, we define a natural-language template describing the physical system, with slots for case-specific physical conditions, parameters, and derived quantities. These slots are instantiated from the simulation metadata, so queries within a family share similar wording but differ in the physical properties and numerical values associated with each case. See Figure~\ref{fig:query-highlight-example} for an example of a natural-language template instantiated with metadata values.

\paragraph{Regression}

To evaluate whether quantitative physical information is recoverable from video representations, we define ten family-specific physical-property regression tasks. Each task associates one simulation family with a continuous scalar target, such as gravity, Reynolds number, wavelength, or elastic modulus. Given frozen video embeddings, we fit a linear regression probe to predict the target values $\hat{\mathbf{y}}=[\hat{y}_1,\ldots,\hat{y}_N]$ from video representations alone. We report normalized root mean squared error (nRMSE) as the primary metric and Spearman's rank correlation $\rho$ as a complementary measure of whether the predicted values preserve the ordering of the ground-truth property. Details of the regression targets and evaluation protocol are provided in Appendix~\ref{app:regression_protocol}.

\paragraph{Pair Classification}

Pair classification provides a lower-cost, controlled test of whether models can distinguish the correct video-text correspondence from alternatives. We sample 10 anchors per family and pair each with a \textit{within-family} negative and a \textit{cross-family} negative, yielding two binary settings. The former tests fine-grained sensitivity to physical properties and numerical values, while the latter tests discrimination across physical systems. For embedding models, we compare the anchor-positive similarity against the corresponding anchor-negative similarity. For MLLMs, we directly prompt the model to answer each binary multiple-choice question, rather than evaluating its embeddings. We report accuracy for both settings.

\section{Experiments}

\subsection{Main Results}

We evaluate three classes of models with different representation paradigms. 
\textbf{Embedding models} include Video-CLIP-XL~\citep{wang2024videoclip}, Qwen3-VL-Embedding (2B and 8B)~\citep{li2026qwen3vl}, Omni-Embed-Nemotron 3B~\citep{xu2025omni}, and LCO-Embedding-Omni (3B and 7B) \citep{xiao2026scaling}. These models provide both video and text embeddings and are therefore evaluated on all three benchmark tasks. \textbf{JEPA models} are represented by V-JEPA2~\citep{assran2025vjepa2selfsupervisedvideo}. As V-JEPA2 produces video representations without a corresponding text embedding space, we evaluate it only on physical-property regression using frozen video features. \textbf{MLLMs} include Qwen3.5 \citep{qwen3.5} and Qwen3.6 \citep{qwen36_35b_a3b, qwen3.6-27b}, each with 27B and 35B-A3B variants. Rather than extracting embeddings, we directly prompt these models to answer the pair-classification questions, providing a generative-model reference for physical video-text discrimination. 

The main results are summarized in Table~\ref{tab:main_results}. We make three observations. First, current pre-trained embedding models remain weak at physics-aware cross-modal alignment: retrieval performance stays low across all four physics branches, and within-family pair classification remains close to random chance, although cross-family discrimination is considerably easier. Second, directly prompted MLLMs perform substantially better on pair classification, reaching around 91--92\% cross-family accuracy and 59--61\% within-family accuracy. 
This suggests that stronger multimodal generative models already possess meaningful capabilities for understanding physical video-text correspondences, whereas current embedding models do not expose comparable capabilities through similarity matching, motivating us to study whether such capabilities can be better activated for representation learning.
Third, V-JEPA2, despite being trained without text alignment, achieves strong physical-property regression performance with frozen video features. This suggests that predictive video representation learning itself can preserve substantial quantitative physical information, highlighting a distinction between modeling physical information in the representation and aligning it with language for similarity matching.

\input{Tables/main_table}

\subsection{Enhancement of Physics Capabilities}

We next ask whether physics adaptation can improve embedding model performance on our benchmark. Existing embedding models are rarely trained explicitly to capture physical information. Among the two dominant paradigms for multimodal retrieval, prior work has identified important mechanistic differences between CLIP and MLLM-based embedding models \citep{xiao2025mieb,xiao2026scaling}. First, CLIP models exhibit a scaling plateau on complex tasks requiring spatial and compositional cross-modal alignment, where simply scaling model size or data provides limited gains. Second, MLLMs acquire richer multimodal knowledge through generative pretraining, as the LLM decoder learns to leverage multimodal information in a shared representation space; lightweight contrastive learning can then activate this latent alignment for similarity matching. Third, a generation-representation scaling pattern emerges: stronger generative models tend to provide higher representation upper bounds after contrastive activation.

These observations motivate us to focus on the MLLM-based paradigm and ask whether physics capabilities already present in generative models can be released through physics-aware contrastive learning. Our setup therefore calls for a backbone with strong world knowledge, a validated contrastive activation recipe that permits injecting physics-aware examples while retaining general-domain data, and omnimodal support for video and audio, which also enables downstream retrieval-augmented video generation.

We adopt the training recipe of LCO-Embedding \citep{xiao2026scaling}, an omnimodal embedding model trained with lightweight contrastive activation that achieves competitive performance across text, image, video, and audio embedding benchmarks.

\begin{figure}[ht]
    \centering
    \includegraphics[width=1.0\linewidth]{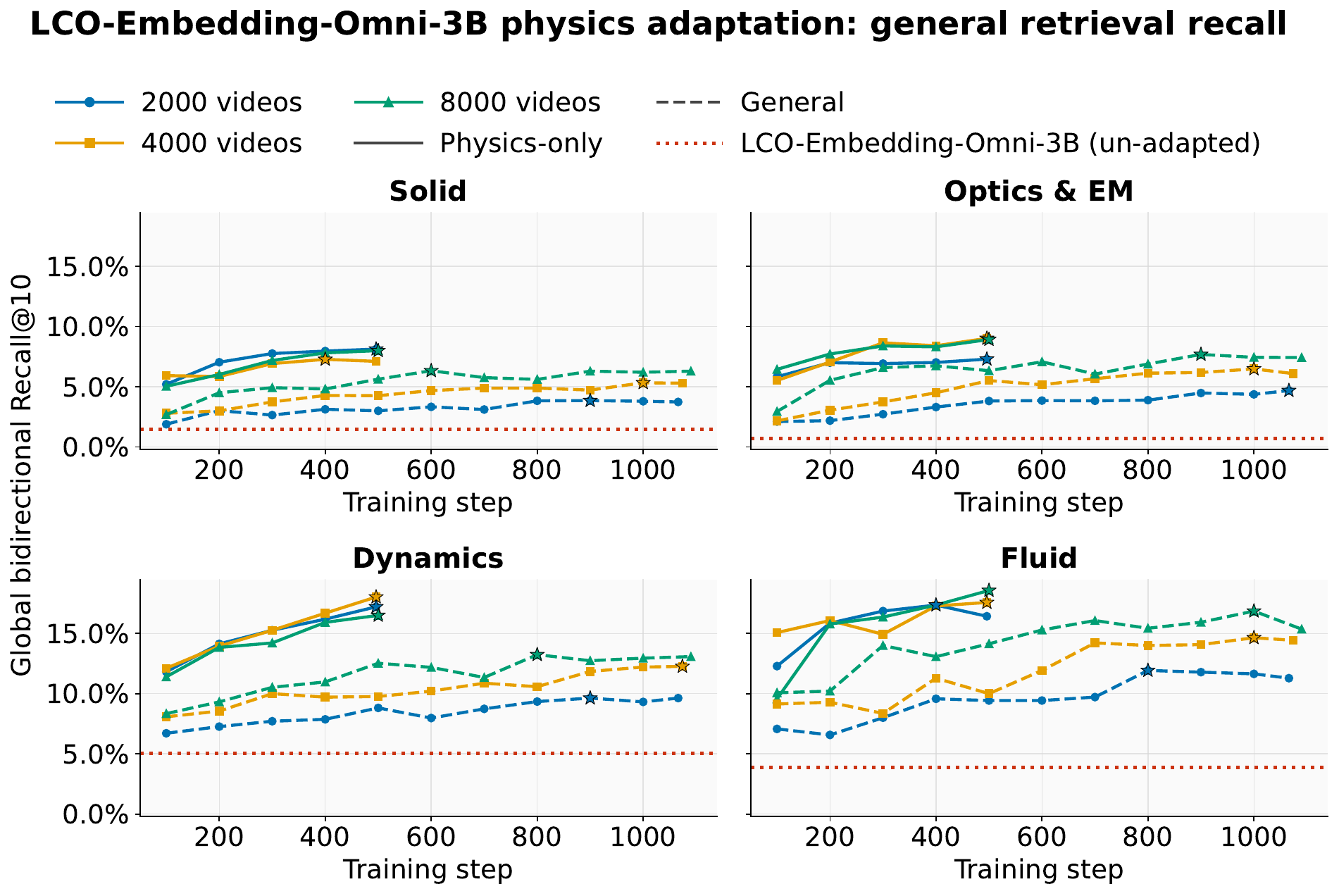}
    \caption{Retrieval performance (y-axis) across training steps (x-axis) and training dataset settings (different lines).}
    \label{fig:retrieval_after_training}
\end{figure}

\paragraph{Experiment Setups} We construct the training set using the same physics simulation engines while ensuring that physics quantities take different numerical values from those in the benchmark. In total, we generate 8,000 text–video pairs by simulating 100 instantiations for each of the 80 physics families. Videos are rendered from the simulation results, with captions expressing the corresponding physics quantities in natural language following the benchmark templates. For each anchor, we sample one hard negative from the same scene family.

We conduct two training settings. \textbf{Physics-only Training} uses only physics video-text pairs, with nested subsets of \{$2$k, $4$k, $8$k\} examples. Given the small dataset size and high visual similarity within physics families, we use a batch size of $64$ and train for \{$16$, $8$, $4$\} epochs, respectively, to keep the number of optimization steps approximately constant. \textbf{General Training} mixes the same physics subsets with $271$k general video-text pairs from the \texttt{LCO-Embedding-Omni-3B-2605} collection, yielding \{$273$k, $275$k, $279$k\} examples. We initially hypothesize that general-domain data can stabilize training and prevent capability collapse, although later results provide evidence against this hypothesis. We use a batch size of $512$, and following \citet{xiao2026scaling}, we train all three settings for $2$ epochs; their total training steps differ only slightly because physics data constitute a small fraction of the mixture.

We initialize from \texttt{LCO-Embedding-Omni-3B} and \texttt{LCO-Embedding-Omni-7B}. We find that initializing directly from \texttt{Qwen2.5-Omni} (which is itself the base model of LCO-Embedding) leads to frequent loss spikes when physics data are introduced. We attribute this to the anisotropic representation space of the generative model, whereas LCO-Embedding has already been calibrated for similarity-based representation learning.

\paragraph{Results and Analyses} Figure~\ref{fig:retrieval_after_training} summarizes retrieval performance of models across training settings and steps. We make two observations: First, contrastive learning substantially improves physics-aware video retrieval for MLLM-based embedding models. Although recall@10 remains well below that of general-domain video-text retrieval \citep{assadi2026mveb}, it clearly surpasses the near-chance performance of the base model in Table~\ref{tab:main_results}. Second, physics-aware retrieval requires prolonged optimization. Retrieval performance continues to improve throughout training and shows no signs of converging within our allocated training resources. For the physics-only training with $2000$ examples, this results in an unusually high number of $16$ epochs. Moreover, mixing general-domain video-text data slows down the training. Both behaviors contrast with general-domain MLLM-based embedding model training, where $\leq 2$ epochs is typically sufficient \citep{xiao2026scaling}.

\begin{wrapfigure}{r}{0.4\columnwidth}
    \centering
    \includegraphics[width=\linewidth]{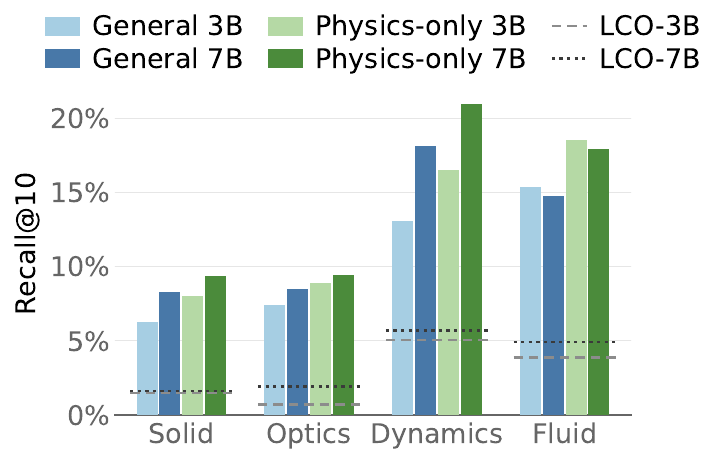}
    \caption{Retrieval performance of 3B and 7B backbones after physics adaptation across the four physics branches.}
    \label{fig:3b_7b_final_checkpoint_bar}
\end{wrapfigure}

Figure~\ref{fig:3b_7b_final_checkpoint_bar} compares checkpoints trained from 3B and 7B backbones to test whether physics retrieval scales with backbone capability. Consistent with \cite{xiao2026scaling,xiao2025mieb}, stronger generative backbones yield stronger retrieval performance after identical contrastive training, supporting the generation-representation scaling law in physics representation learning. Figure~\ref{fig:retrieval_after_training_7b} displays the retrieval performance of 7B models throughout the physics adaptation training.

Regression performance in Figure~\ref{fig:regression_after_training}, however, calls for a more nuanced interpretation. Across the 10 regression tasks, models continually trained on in-domain physics data mostly, if not consistently, underperform \texttt{LCO-Embedding-Omni}. We provide two possible interpretations. First, this observation is broadly consistent with \cite{xiao2026scaling}, which argues that contrastive learning primarily activates and calibrates MLLM capabilities rather than enhancing the underlying capabilities themselves. While in-domain physics training improves retrieval by better calibrating the similarity-matching space, a linear regression probe instead measures whether specific information is encoded and linearly accessible in the embeddings. As a side effect of this representation calibration, the model's ability to encode certain physics properties may therefore degrade. Second, physics-property regression may benefit from training objectives beyond contrastive learning, such as explicit representation regression losses, similar to objectives used to optimize semantic textual similarity in text embedding models.

\subsection{Retrieval-Augmented Video Generation}

Having attained stronger video-text alignment in the physics representation space, we analyze whether such physics-aware representation can enhance video generation. We consider retrieval-augmented video generation. We equip MiniMax-H3 \citep{minimaxh32026}, a state-of-the-art open-weight video generation model, with multimodal embedding models serving as retrievers. We leverage the external video database built in \citep{cheng2026physrag}, and use the \texttt{ref2va} version of Minimax-H3 \citep{minimaxh32026}, which inherently has certain capabilities of grounding on external contexts.

We benchmark the performance on PhyGenBench~\citep{meng2025towards}, and take a closer look at two physics branches related to World Embedding Benchmark. The results are displayed in Table~\ref{tab:rag_generation}. As a frontier open-weight model, MiniMax-H3 without RAG already achieves strong PhyGenBench performance. When forced to refer to potentially suboptimal videos, average performance slightly degrades from the baseline, shown in the VideoCLIP-XL \citep{wang2024videoclip} column. The original LCO-Embedding-Omni-3B \citep{xiao2026scaling} enhances the performance over the no-RAG baseline. The LCO variant adapted for physics-aware video retrieval in this work further improves performance, with the \textit{general + physics} variant attaining the best average score of 0.66 and the physics-only variant reaching 0.65.

\begin{table}[t]
\centering
\small
\setlength{\tabcolsep}{5pt}
\renewcommand{\arraystretch}{1.1}
\caption{Retrieval-augmented video generation with MiniMax-H3 using top-1 retrieved references.}
\label{tab:rag_generation}
\vspace{4pt}

\resizebox{0.9\linewidth}{!}{%
\begin{tabular}{lccccc}
\toprule
& \makecell[c]{No RAG}
& \makecell[c]{VideoCLIP-XL}
& \makecell[c]{LCO-Omni-3B}
& \makecell[c]{LCO-Omni-3B + Ft.\\(general+physics)}
& \makecell[c]{LCO-Omni-3B + Ft.\\(physics-only)} \\
\midrule

Optics
& 0.67 & 0.63 & 0.65 & 0.68 & 0.65 \\

Thermal
& 0.53 & 0.54 & 0.61 & 0.63 & 0.64 \\

\midrule
Average
& 0.60 & 0.59 & 0.63 & \textbf{0.66} & \underline{0.65} \\

\bottomrule
\end{tabular}%
}
\end{table}

\section{Related Work}

\subsection{Physics Simulation}

Numerical benchmarks evaluate surrogate models on parameterized physical systems \citep{takamoto2022pdebench,bonnet2022airfrans,gupta2023towards,toshev2023lagrangebench,luo2024cfdbench,ohana2024well}, while visual benchmarks study causal reasoning and event prediction in simulated environments \citep{riochet2020intphys,yi2020clevrer,bear2021physion,chen2025comphy}. Existing work therefore focuses mainly on either numerical fields or object-level dynamics. Our benchmark instead renders simulations across four physics domains as videos paired with physical quantities, enabling retrieval and physical property regression on video representations.

\subsection{Physics-aware Video Generation and World Modeling}

Video generation evaluation has expanded from visual quality and text alignment \citep{liu2023fetv,liu2024evalcrafter,huang2024vbench} to physical plausibility and world-modeling capabilities \citep{bansal2025videophy,meng2025towards,zheng2025vbench2, duan2025worldscore,li2026worldmodelbench}. Related methods improve physical faithfulness through prompt refinement \citep{xue2025phyt2v}, physics-oriented training \citep{wang2025wisa}, or retrieved video references \citep{cheng2026physrag}.  
In contrast, we examine the physical information encoded in pretrained video representations and its utility for retrieval, regression, and retrieval-augmented generation.

\subsection{Omnimodal Representation Learning}

Omnimodal representation learning has progressed from image-text alignment \citep{radford2021clip} to unified embedding spaces across modalities \citep{girdhar2023imagebind,zhu2024languagebind}. Recent MLLM-based approaches obtain general-purpose embeddings through text-only contrastive learning \citep{jiang2024e5v}, instruction-conditioned training \citep{jiang2025vlm2vec}, or language-centric contrastive refinement \citep{xiao2026scaling} to activate pretrained capabilities \citep{xiao2024rar}. Recent models support videos and visual documents \citep{meng2026vlmvecv,li2026qwen3vl,xiao2024pixel,yuan2026design}, interleaved inputs \citep{zhou2026wemm}, and audio or composed multimodal queries \citep{huynh2026omniret}. \cite{xiao2025mieb,assadi2026maeb,assadi2026mveb,zhang2026reasonaudio,huang2026mmebv} have expanded the evaluation, covering a broader range of modalities and reveals persistent modality bias and asymmetric cross-modal retrieval. Complementing this broad semantic evaluation, our benchmark focuses on whether video embeddings preserve fine-grained physical information, as measured by physical-description alignment and quantitative property recovery.

\section{Conclusion}
We introduced the World Embedding Benchmark to evaluate how video representations capture and align physical information. Across 8,000 controlled simulation cases, we find that pre-trained omnimodal embeddings preserve recoverable physical signals but exhibit weak cross-modal physical alignment. Physics-specific contrastive adaptation improves retrieval and pair classification, but degrades physical-property regression, revealing a trade-off between alignment and quantitative information recoverability. Finally, physically relevant retrieval improves the fidelity of generated videos, with stronger retrieval models yielding larger gains. These results highlight the importance of jointly evaluating physical alignment and information recoverability when developing representations for physically grounded world models.

\bibliography{iclr2027_conference}
\bibliographystyle{iclr2027_conference}

\newpage

\appendix

\section{Authors}
\label{app:authors}

Yiqi Liu* \quad The University of Manchester

Ruifeng Yuan* \quad The Hong Kong Polytechnic University

Yang Wang* \quad The University of Manchester

Long Li \quad Zhejiang University

Fengyu Cai \quad Technical University of Darmstadt

Hou Pong Chan \quad University of Macau

Jialin Yu \quad University of Oxford

Hao Zhang \quad Nanyang Technological University

Chenghua Lin \quad The University of Manchester

Chenghao Xiao\textdagger \quad Shanghai University of Finance and Economics

* Core Contribution \quad \textdagger Corresponding Author

\section{Dataset and Generation Details}
\label{app:A}

\subsection{Dataset Overview}
\label{app:A1}

The physics simulation dataset contains 8,000 cases organized into 80 families across four branches: Fluid Mechanics, Solid Mechanics, Dynamics, and Optics and Electromagnetism. Each family contains 100 cases. Cases within a family share the same physical scenario while varying selected material properties, geometric parameters, initial or boundary conditions, loading conditions, or excitation parameters.

Each case contains its simulation configuration, computed physical response, associated physical quantities, and rendered video. These records provide the source data for the retrieval, physical-property regression, and pair-classification tasks described in Appendix~\ref{app:evaluation_protocols}.

\subsection{Physical Formulations and Simulation}
\label{app:A2}

The simulation families use established simulation backends, analytical evaluation, or numerical ODE integration according to the underlying physical formulation. In this appendix, \emph{Analytical / ODE integration} denotes analytical evaluation or numerical time integration implemented directly from the specified formulation. The formulations below summarize the equations used by the generation pipelines. Table~\ref{tab:family_physics_mapping} maps each family to its formulation, computation method, case parameters, and field output.

\subsubsection{Fluid Mechanics}
\label{app:A2-fluid}

\paragraph{Governing formulations}
The fluid families use the incompressible mass and momentum equations,
\begin{equation}
\label{eq:fluid-ns}
\begin{aligned}
\nabla\!\cdot\!\mathbf{u} &= 0,\\
\frac{\partial\mathbf{u}}{\partial t}
+(\mathbf{u}\!\cdot\!\nabla)\mathbf{u}
&=
-\frac{1}{\rho}\nabla p
+\nabla\!\cdot\!\left(2\nu_{\mathrm{eff}}\mathbf{D}\right)
+\mathbf{f},\\
\mathbf{D}
&=
\frac{1}{2}
\left(\nabla\mathbf{u}+\nabla\mathbf{u}^{T}\right).
\end{aligned}
\end{equation} Here, $\nu_{\mathrm{eff}}$ is the kinematic viscosity used by the selected laminar, turbulence, or rheological configuration.

\paragraph{Family organization}
The fluid subset contains seven benchmark families instantiated from fifteen lower-level simulation configurations. The family organization groups configurations according to their dominant physical mechanism instead of their geometry. These families cover internal flows, external flows, transport phenomena, buoyancy-driven flows, and non-Newtonian rheology.

\paragraph{Family-specific relations}
Buoyancy-driven configurations couple thermal transport to momentum through the Boussinesq approximation,
\begin{equation}
\label{eq:fluid-boussinesq}
\begin{aligned}
\frac{\partial T}{\partial t}
+\mathbf{u}\!\cdot\!\nabla T
&=\alpha\nabla^2T,\\
\mathbf{f}_{b}
&=\mathbf{g}\,\beta(T-T_0).
\end{aligned}
\end{equation} The non-Newtonian configurations use a generalized-Newtonian stress relation,
\begin{equation}
\label{eq:fluid-rheology}
\boldsymbol{\tau}
=
2\rho\nu_{\mathrm{eff}}(\dot{\gamma})\mathbf{D},
\qquad
\dot{\gamma}
=
\sqrt{2\mathbf{D}:\mathbf{D}},
\end{equation}
where $\nu_{\mathrm{eff}}$ follows the PowerLaw or HerschelBulkley transport model selected by the OpenFOAM case. The corresponding case configuration specifies $K$, $n$, $\tau_0$, $\nu_0$, $\nu_{\min}$, and $\nu_{\max}$ as applicable. Passive-scalar transport follows
\begin{equation}
\label{eq:fluid-scalar}
\frac{\partial c}{\partial t}
+\mathbf{u}\!\cdot\!\nabla c
=
D\nabla^2c.
\end{equation}

\paragraph{Computation and visualization}
Fluid cases are solved with OpenFOAM using the finite-volume method. The principal field outputs are velocity magnitude and, for the corresponding families, temperature or transported-scalar fields. Video frames follow simulation time.

\subsubsection{Solid Mechanics}
\label{app:A2-solid}

\paragraph{Governing formulations}
The linear-elastic families solve static equilibrium with the small-strain isotropic constitutive relation,
\begin{equation}
\label{eq:solid-linear}
\begin{aligned}
\nabla\!\cdot\!\boldsymbol{\sigma}+\mathbf{b}
&=\mathbf{0},\\
\boldsymbol{\varepsilon}
&=
\frac{1}{2}
\left(\nabla\mathbf{u}+\nabla\mathbf{u}^{T}\right),\\
\boldsymbol{\sigma}
&=
\lambda\,\mathrm{tr}(\boldsymbol{\varepsilon})\mathbf{I}
+2\mu\boldsymbol{\varepsilon}.
\end{aligned}
\end{equation}

Finite-strain families use the compressible Neo-Hookean energy implemented by the solver,
\begin{equation}
\label{eq:solid-finite-strain}
\begin{aligned}
\mathbf{F}
&=\mathbf{I}+\nabla\mathbf{u},
&
J&=\det\mathbf{F},
&
\mathbf{C}&=\mathbf{F}^{T}\mathbf{F},\\
\psi(\mathbf{F})
&=
\frac{\mu}{2}
\left[
\mathrm{tr}(\mathbf{C})-d-2\ln J
\right]
+
\frac{\kappa}{2}
(\ln J)^2,\\
\operatorname{Div}\mathbf{P}
&=
\mathbf{0},
&
\mathbf{P}
&=
\frac{\partial\psi}{\partial\mathbf{F}}.
\end{aligned}
\end{equation} In~\eqref{eq:solid-finite-strain}, $\kappa$ denotes the volumetric coefficient multiplying $(\ln J)^2$. The geometrically nonlinear buckling family uses the same finite-strain energy form together with an initial geometric imperfection and compressive loading.

\paragraph{History-dependent response}
Families carrying J2-related identifiers use the internal-variable update implemented by the current DOLFINx pipeline,
\begin{equation}
\label{eq:solid-history}
\begin{aligned}
\varepsilon_{\mathrm{eq}}
&=
\sqrt{
\frac{2}{3}
\,\mathrm{dev}(\boldsymbol{\varepsilon})
:\mathrm{dev}(\boldsymbol{\varepsilon})
},\\
p^{k+1}
&=
p^k
+c_p
\left\langle
\varepsilon_{\mathrm{eq}}
-
\left(
c_y\varepsilon_y
+c_H\frac{H}{E}p^k
\right)
\right\rangle_{+},\\
\mu_{\mathrm{eff}}
&=
\frac{\mu}{1+\beta p}.
\end{aligned}
\end{equation} The coefficients $c_p$, $c_y$, and $c_H$ follow the corresponding 2D or 3D implementation and the selected loading family. The~\eqref{eq:solid-history} is the implemented history-dependent constitutive update used to generate these cases.

\paragraph{Computation and visualization}
Solid-mechanics cases are solved with DOLFINx using finite-element discretization. The released videos primarily visualize displacement magnitude on the deformed geometry. Frame progression follows successive loading states.

\subsubsection{Dynamics}
\label{app:A2-dynamics}

\paragraph{Rigid-body and lumped-parameter dynamics}
Chrono-based families use constrained Newton-Euler dynamics,
\begin{equation}
\label{eq:dyn-rigid}
\begin{aligned}
m_i\dot{\mathbf{v}}_i
&=\sum\mathbf{F}_i,\\
\mathbf{I}_i\dot{\boldsymbol{\omega}}_i
+\boldsymbol{\omega}_i\times
(\mathbf{I}_i\boldsymbol{\omega}_i)
&=\sum\boldsymbol{\tau}_i,\\
\boldsymbol{\Phi}(\mathbf{q},t)
&=\mathbf{0}.
\end{aligned}
\end{equation} 

Rigid collision uses an impulse-based contact update. The normal component satisfies
\begin{equation}
\label{eq:dyn-impact}
m_i
\left(
\mathbf{v}_i^{+}-\mathbf{v}_i^{-}
\right)
=
\pm J_n\mathbf{n},
\qquad
\mathbf{n}\cdot\mathbf{v}_{\mathrm{rel}}^{+}
=
-e\,
\mathbf{n}\cdot\mathbf{v}_{\mathrm{rel}}^{-}.
\end{equation}
The implementation additionally accounts for eccentric contact and the associated rotational response, together with a bounded tangential friction impulse when friction is enabled.

The mass-spring-damper family integrates
\begin{equation}
\label{eq:dyn-msd}
\mathbf{M}\ddot{\mathbf{q}}
+\mathbf{C}\dot{\mathbf{q}}
+\mathbf{K}\mathbf{q}
=
\mathbf{f}(t),
\end{equation}
and the rolling/sliding family uses
\begin{equation}
\label{eq:dyn-rolling}
m\dot{\mathbf{v}}=\sum\mathbf{F},
\qquad
I\dot{\omega}=\sum\tau,
\qquad
v_{\mathrm{rel}}=v-R\omega.
\end{equation}
Drop-weight impact is integrated using a spring-damper contact force coupled to an elastic plate degree of freedom,
\begin{subequations}
\label{eq:dyn-contact}
\begin{gather}
\delta
=
\left[z_{\mathrm{imp}} + r - z_{\mathrm{plate}}\right]_{+},
\label{eq:dyn-contact-gap}
\\
F_c
=
\left[
k_n \delta
+ c_n \left(v_{\mathrm{imp}} - v_{\mathrm{plate}}\right)
\right]_{+},
\label{eq:dyn-contact-force}
\\
m_{\mathrm{plate}} \ddot q
+ c_{\mathrm{plate}} \dot q
+ k_{\mathrm{plate}} q
=
F_c.
\label{eq:dyn-contact-response}
\end{gather}
\end{subequations}

\paragraph{Mechanism relations}
The crank-slider family uses geometric closure,
\begin{equation}
\label{eq:dyn-crank}
\theta(t)=\theta_0+\omega t,
\qquad
x_s
=
r\cos\theta
+\sqrt{L^2-r^2\sin^2\theta}.
\end{equation}
The gear family uses tooth-count kinematics,
\begin{equation}
\label{eq:dyn-gear}
\omega_j
=
-\frac{N_i}{N_j}\omega_i,
\qquad
\theta_j-\theta_{j,0}
=
-\frac{N_i}{N_j}
(\theta_i-\theta_{i,0}).
\end{equation}
The articulated-chain generator uses a damped, driven angular response,
\begin{equation}
\label{eq:dyn-chain}
\theta_j(t)
=
\theta_{0,j}e^{-\zeta_j t}
\cos(\omega_j t+\phi_j)
+
A_d r_d^{\,j}\sin(\Omega t-j\Delta\phi)
+
c_j\theta_{j-1}(t),
\end{equation}
with $\omega_j$ determined from gravity and the mean upstream link length in the implementation.

\paragraph{Structural-response relations}
The transient cantilever family uses the Euler-Bernoulli tip displacement and first-mode frequency,
\begin{equation}
\label{eq:dyn-cantilever}
\delta_s
=
\frac{FL^3}{3EI},
\qquad
f_1
=
\frac{1.875^2}{2\pi}
\sqrt{\frac{EI}{\rho A L^4}}.
\end{equation}
The harmonic-response family uses
\begin{equation}
\label{eq:dyn-frf}
A(\omega)
=
\frac{F_0}
{\sqrt{(k-m_{\mathrm{eff}}\omega^2)^2+(c_{\mathrm{eff}}\omega)^2}}.
\end{equation}
The dataset identifier \texttt{reissnerShellModal} currently uses a thin-plate modal generator,
\begin{equation}
\label{eq:dyn-plate}
D
=
\frac{Et^3}{12(1-\nu^2)},
\qquad
w_{mn}(x,y,t)
=
A_{mn}
\sin\frac{m\pi x}{a}
\sin\frac{n\pi y}{b}
\sin(2\pi f_{mn}t)e^{-0.12t}.
\end{equation}
Longitudinal stress-wave propagation follows
\begin{equation}
\label{eq:dyn-wave}
\frac{\partial^2u}{\partial t^2}
=
c^2\frac{\partial^2u}{\partial x^2},
\qquad
c=\sqrt{\frac{E}{\rho}}.
\end{equation}
The unbalanced-rotor response uses the rotating-imbalance amplitude relation
\begin{equation}
\label{eq:dyn-rotor}
r=\frac{\Omega}{\Omega_c},
\qquad
A(r)
=
e\,
\frac{r^2}
{\sqrt{(1-r^2)^2+(2\zeta r)^2}}.
\end{equation}

\paragraph{Parameterized structural responses}
The \texttt{j2PlasticBeam}, \texttt{lemaitreDamageBar}, and \texttt{rubberBlockCompression} families use explicit parameterized response relations defined in their generation pipelines. Their displacement, internal-variable, and deformation fields are given by~\eqref{eq:dyn-plastic-beam}--\eqref{eq:dyn-rubber-compression}.

\begin{subequations}
\label{eq:dyn-prescribed}

\begin{equation}
\label{eq:dyn-plastic-beam}
u_{\mathrm{beam}}(x,t)
=
a(t)\phi(x),
\qquad
\bar{\varepsilon}^{p}(x,t)
=
\left(1-\frac{x}{L}\right)^2 p(t).
\end{equation}

\begin{equation}
\label{eq:dyn-damage-bar}
D(x,t)
=
d_{\max}(t)
\exp\!\left[
-\frac{(x-L/2)^2}{2w^2}
\right],
\qquad
u_{\mathrm{damage}}(x,t)
=
u_0\sin(2\pi f t)\frac{x}{L}
\left[1-0.55D(x,t)\right].
\end{equation}

\begin{equation}
\label{eq:dyn-rubber-compression}
x'
=
x\left[1+0.24c(t)\right],
\qquad
y'
=
y\left[1+0.24c(t)\right],
\qquad
z'
=
z\left[1-c(t)\right].
\end{equation}

\end{subequations}

Here, $a(t)$ and $p(t)$ specify the prescribed beam-response histories, $d_{\max}(t)$ controls the evolution of the localized damage field, and $c(t)$ specifies the compression history of the rubber-block family. The spatial functions $\phi(x)$ and the Gaussian profile in~\eqref{eq:dyn-damage-bar} determine the corresponding field distributions.

\paragraph{Computation and visualization}
Dynamics combines Chrono and Analytical / ODE integration of the specified equations. The displayed quantities depend on the family and include velocity magnitude, displacement, angular speed, axial stress, equivalent plastic strain, and scalar damage. Frames represent physical time.

\subsubsection{Optics and Electromagnetism}
\label{app:A2-optics}

\paragraph{Geometric optics}
Reflection, paraxial imaging, and refraction use
\begin{equation}
\label{eq:opt-reflection}
\mathbf{d}_r
=
\mathbf{d}_i
-2(\mathbf{d}_i\!\cdot\!\mathbf{n})\mathbf{n},
\end{equation}
\begin{equation}
\label{eq:opt-imaging}
\frac{1}{f}
=
\frac{1}{d_o}
+\frac{1}{d_i},
\end{equation}
and
\begin{equation}
\label{eq:opt-refraction}
n_1\sin\theta_1
=
n_2\sin\theta_2,
\qquad
\theta_c
=
\sin^{-1}\!\left(\frac{n_2}{n_1}\right),
\quad n_1>n_2.
\end{equation}
Interface reflection uses the Fresnel coefficients,
\begin{equation}
\label{eq:opt-fresnel}
\begin{aligned}
r_s
&=
\frac{n_1\cos\theta_i-n_2\cos\theta_t}
     {n_1\cos\theta_i+n_2\cos\theta_t},\\
r_p
&=
\frac{n_2\cos\theta_i-n_1\cos\theta_t}
     {n_2\cos\theta_i+n_1\cos\theta_t},
\qquad
R=|r|^2.
\end{aligned}
\end{equation}

\paragraph{Wave and polarization optics}
HCIPy families propagate a scalar optical field,
\begin{equation}
\label{eq:opt-propagation}
U_z
=
\mathcal{P}_z\{U_0\},
\qquad
I(x,y)
=
|U_z(x,y)|^2.
\end{equation}
Analytical interference families use
\begin{equation}
\label{eq:opt-interference}
d\sin\theta_m
=
m\lambda,
\qquad
\Delta\phi_{\mathrm{film}}
=
\frac{4\pi nt\cos\theta_t}{\lambda}.
\end{equation}
Gaussian-beam propagation follows
\begin{equation}
\label{eq:opt-gaussian}
w(z)
=
w_0
\sqrt{1+\left(\frac{z}{z_R}\right)^2},
\qquad
z_R
=
\frac{\pi w_0^2}{\lambda}.
\end{equation}
Polarization families use
\begin{equation}
\label{eq:opt-polarization}
I
=
I_0\cos^2\theta,
\qquad
\mathbf{E}_{\mathrm{out}}
=
\mathbf{R}(\theta)
\begin{bmatrix}
1&0\\
0&e^{i\delta}
\end{bmatrix}
\mathbf{R}(-\theta)\mathbf{E}_{\mathrm{in}}. \end{equation}

\paragraph{Electromagnetism}
The Meep families solve Maxwell's equations in the time domain,
\begin{equation}
\label{eq:em-maxwell}
\nabla\times\mathbf{E}
=
-\mu\frac{\partial\mathbf{H}}{\partial t},
\qquad
\nabla\times\mathbf{H}
=
\mathbf{J}
+\epsilon\frac{\partial\mathbf{E}}{\partial t}.
\end{equation}
The plasmonic-dimer family additionally uses the configured Drude-type dispersive material relation,
\begin{equation}
\label{eq:em-drude}
\epsilon(\omega)
=
\epsilon_{\infty}
-
\frac{\omega_p^2}
{\omega^2+i\gamma\omega}.
\end{equation}

\paragraph{Computation and visualization}
Optical families use Analytical / ODE integration or scalar wave propagation with HCIPy. Electromagnetic families use Meep FDTD. Optical field outputs are rendered as irradiance. Meep frames use
\begin{equation}
\label{eq:em-visualization}
Q
=
(\mathrm{Re}\,E_x)^2
+(\mathrm{Re}\,E_y)^2
+(\mathrm{Re}\,E_z)^2
+0.15(\mathrm{Re}\,H_z)^2,
\end{equation}
followed by per-case normalization over the sampled frames. Optical sequences may follow a parameter scan; Meep sequences follow simulation time.

\begingroup
\input{Tables/appendix_family_mapping_table}

\endgroup

\section{Evaluation Protocols}
\label{app:evaluation_protocols}

\subsection{Retrieval}
\label{app:retrieval_protocol}

\paragraph{Queries construction}
The retrieval evaluation contains one primary query for each simulation case. Each simulation case is associated with one primary retrieval query generated deterministically from a family-specific natural-language template. The template combines a description of the physical scenario, the underlying physical mechanism, the visualized quantity, case-specific physical conditions, and quantitative diagnostics. Case-dependent slots are instantiated from the simulation record and derived diagnostics. Figure~\ref{fig:query-highlight-example} illustrates the resulting structure using a representative Dynamics example.

Each query is associated with one designated target video from the same simulation case. During dataset export, duplicate query texts are checked and disambiguated when necessary to preserve a unique primary query for each released case.

\paragraph{Hard-negative sampling} The training and adaptation datasets additionally provide five hard-negative cases for each anchor. For an anchor case, the candidate negatives are the other 99 cases in the same family. Five distinct cases are sampled with a fixed random seed, and the corresponding query texts and videos are kept aligned. The negative assignments are therefore deterministic for a given dataset revision. These training negatives are stored separately from the retrieval evaluation candidate set.

\input{Figures/query_example}

\subsection{Physical Property Regression}
\label{app:regression_protocol}

\paragraph{Task definition}
The physical-property regression benchmark evaluates whether continuous physical quantities can be recovered from frozen video representations. Each selected simulation family defines an independent regression task with one scalar target quantity and physical unit. The regression model receives only the video representation as input; no textual query or description is provided. The ten regression families and their target quantities are summarized in Table~\ref{tab:regression_targets}.

\paragraph{Representation probe}
Given a video $v_i$, the evaluated encoder produces a frozen representation $\mathbf{z}_i$. A linear regression probe is fitted independently for each family to predict the corresponding physical quantity,
\begin{equation}
    \hat{y}_i = f_{\theta}(\mathbf{z}_i).
\end{equation}
The video encoder remains fixed throughout probe training, so the probe measures the accessibility of the target quantity in the learned representation.

\paragraph{Evaluation metrics.}
For each regression task, we compare the predicted values
$\hat{\mathbf{y}}=[\hat{y}_1,\ldots,\hat{y}_N]$
with the ground-truth values
$\mathbf{y}=[y_1,\ldots,y_N]$.
We report normalized root mean squared error (nRMSE) as the primary metric and Spearman's rank correlation $\rho$ as a complementary measure of whether the predicted values preserve the ordering of the underlying physical quantity.

\input{Tables/appendix_regression_targets}

\section{Regression and Pair Classification Results after training}

\begin{figure}[ht]
    \centering
    \includegraphics[width=1.0\linewidth]{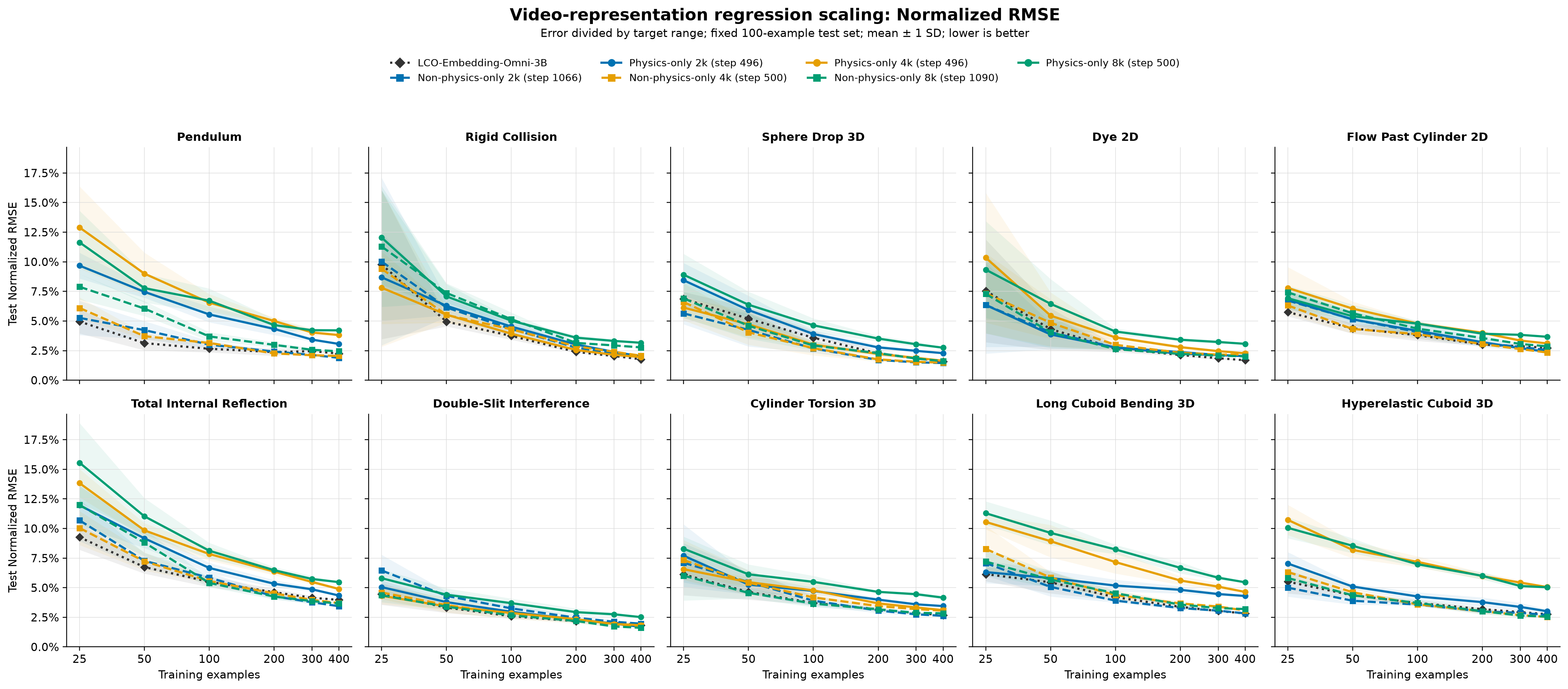}
    \caption{Linear regression performance (y-axis) of different contrastively trained checkpoints (different lines), across different numbers of probe examples (x-axis).}
    \label{fig:regression_after_training}
\end{figure}

\begin{figure}
    \centering
    \includegraphics[width=1.0\linewidth]{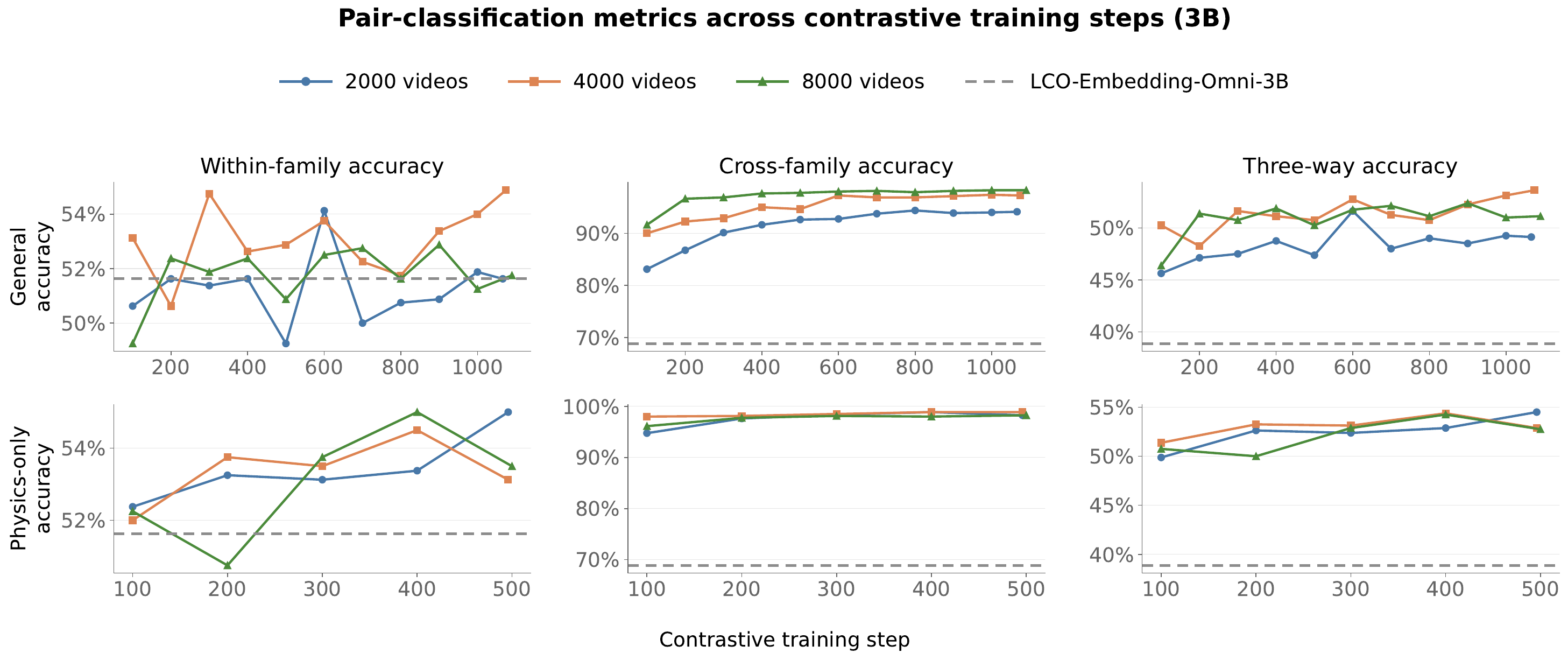}
    \caption{Within-family and cross-family pair classification performance across training steps.}
    \label{fig:pair_classification}
\end{figure}

Figure~\ref{fig:regression_after_training} and Figure~\ref{fig:pair_classification} display results of regression and pair classification across training steps.

\section{Regression and Pair Classification Results after training}
\label{app:scaling_7b_across_training_steps}

\begin{figure}[ht]
    \centering
    \includegraphics[width=1.0\linewidth]{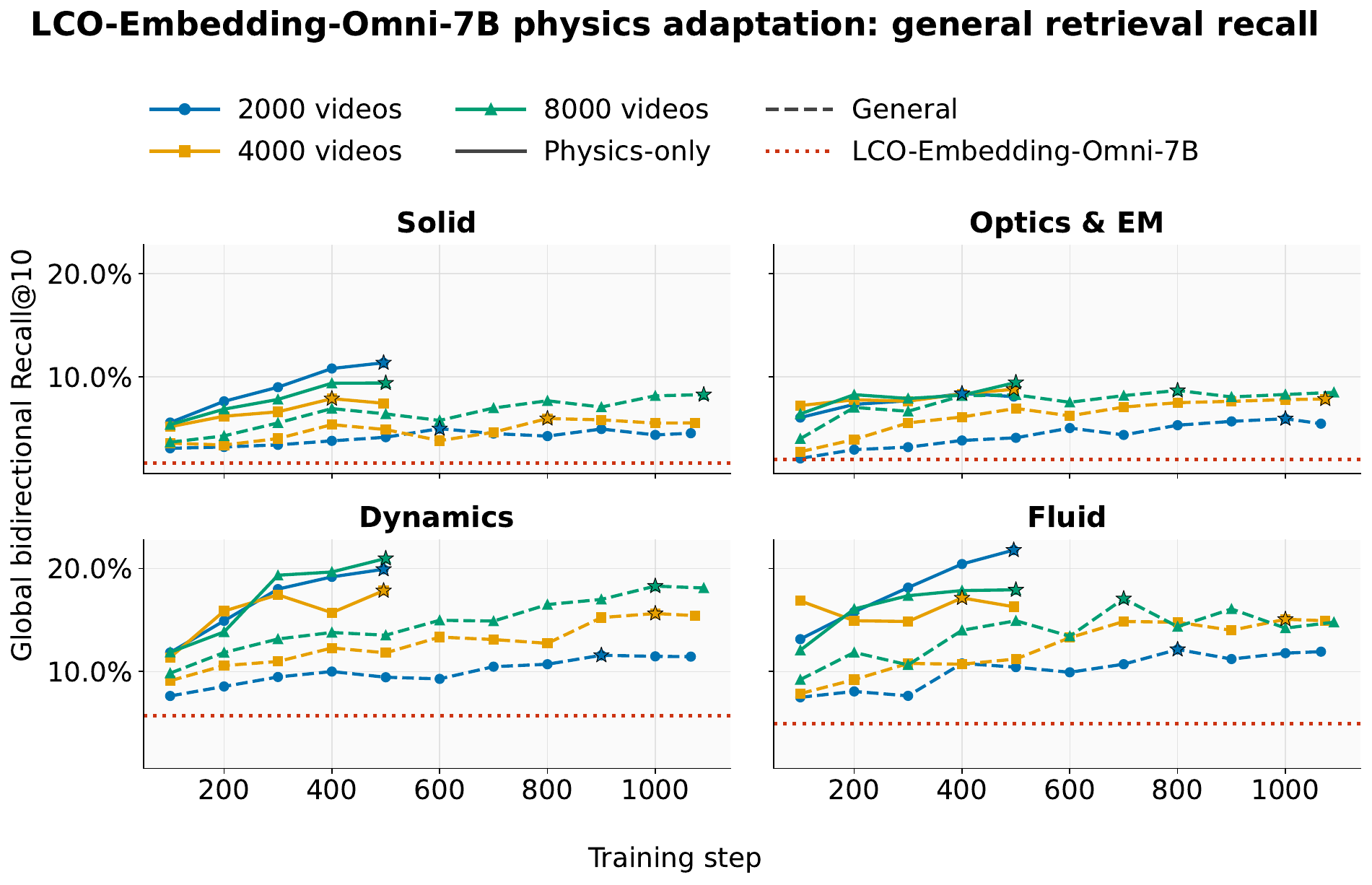}
    \caption{Retrieval performance (y-axis) across training steps (x-axis) and training dataset settings (different lines).}
    \label{fig:retrieval_after_training_7b}
\end{figure}

Figure~\ref{fig:retrieval_after_training_7b} shows the training dynamics from the 7B LCO variant, showing consistent behaviors with the 3B experiments in the main paper.

\end{document}

%% file: math_commands.tex
\usepackage{amsmath,amsfonts,bm}

\def\eqref#1{equation~\ref{#1}}
\def\1{\bm{1}}

\DeclareMathAlphabet{\mathsfit}{\encodingdefault}{\sfdefault}{m}{sl}
\SetMathAlphabet{\mathsfit}{bold}{\encodingdefault}{\sfdefault}{bx}{n}

%% file: Tables/physics_dataset_overview.tex
\begin{table}[htbp] 
\caption{Simulation families, case counts, and computation methods across physics branches.} 
\label{tab:physics_dataset_overview} 
\vspace{4pt}
\centering 
\small 
\resizebox{.65\linewidth}{!}{
\begin{tabular}{lccc} 
\toprule 
Branch & Families & Cases & Computation \\ 
\midrule 
Fluid Mechanics & 7 & 700 & OpenFOAM \\ 
Solid Mechanics & 27 & 2,700 & DOLFINx \\ 
Dynamics & 19 & 1,900 & Chrono + Analytical / ODE integration \\ 
Optics \& EM & 27 & 2,700 & Analytical + HCIPy + Meep (FDTD) \\ 

\midrule 
Total & 80 & 8,000 & -- \\ 
\bottomrule \end{tabular} }
\end{table}

%% file: Tables/main_table.tex
\begin{table*}[t]
\centering
\small
\setlength{\tabcolsep}{4.5pt}
\renewcommand{\arraystretch}{1.12}
\caption{
Performance on the World Embedding Benchmark.
Retrieval is evaluated globally within each physics branch.
Regression reports the macro-average nRMSE across 10 physical-property regression tasks.
Pair classification reports within-family and cross-family accuracy. \textbf{Bold} denotes best across all models, \underline{underline} denotes second-best; * denotes best in embedding models.
}
\label{tab:main_results}
\vspace{4pt}
\resizebox{\textwidth}{!}{%
\begin{tabular}{llccccccc}
\toprule
\multirow{2}{*}{\makecell[c]{Model}}
& \multirow{2}{*}{\makecell[c]{Modality}}
& \multicolumn{4}{c}{\makecell[c]{Retrieval R@10 $\uparrow$}}
& \multicolumn{1}{c}{\makecell[c]{Regression nRMSE $\downarrow$}}
& \multicolumn{2}{c}{\makecell[c]{Pair Classification Acc. $\uparrow$}} \\
\cmidrule(lr){3-6}
\cmidrule(lr){7-7}
\cmidrule(lr){8-9}
& & \makecell[c]{Fluid}
& \makecell[c]{Solid}
& \makecell[c]{Dynamics}
& \makecell[c]{Optics \& EM}
& \makecell[c]{Avg. ($\times 100$)}
& \makecell[c]{Within-Fam.}
& \makecell[c]{Cross-Fam.} \\
\midrule

\multicolumn{9}{l}{\textit{Embedding Models}} \\

Video-CLIP-XL
& T, V & 3.3 & 1.4 & 2.3 & 0.6 & \textbf{0.81} & 48.1 & 66.1 \\

Qwen3-VL-Embedding-2B
& T, I, V & 3.9 & 1.6 & 7.3 & 2.3 & 1.76 & 51.5 & 80.8 \\

Qwen3-VL-Embedding-8B
& T, I, V & 6.6 & 1.9 & 7.4 & 3.1 & 1.80 & 53.1 & 86.2 \\

Omni-Embed-Nemotron-3B
& T, I, V, A & 3.1 & 1.4 & 3.1 & 0.7 & 1.41 & 52.5 & 71.2 \\

% Ovis-Omni-Embedding-3B
% & T, I, V, A & 3.5 & 1.5 & 4.5 & 0.7 & 1.71 & 47.5 & 71.1 \\

LCO-Embedding-3B
& T, I, V, A & 3.9 & 1.5 & 5.1 & 0.7 & 2.47 & 51.5 & 68.5 \\

LCO-Embedding-3B-2605
& T, I, V, A & 4.6 & 1.5 & 4.5 & 1.2 & 1.85 & 50.5 & 72.9 \\

LCO-Embedding-7B
& T, I, V, A & 4.9 & 1.6 & 5.7 & 1.9 & 2.02 & 46.8 & 76.4 \\

\midrule

\multicolumn{9}{l}{\textit{Physics-adapted Embedding Models}} \\

LCO-Embedding-3B + Physics Adaptation
& T, I, V, A & \textbf{18.6} & \underline{8.0} & \underline{16.5} & \underline{8.9} & 3.95 & 53.5 & \underline{98.2} \\

LCO-Embedding-7B + Physics Adaptation
& T, I, V, A & \underline{17.9} & \textbf{9.4} & \textbf{21.0} & \textbf{9.4} & 2.89 & 59.6* & \textbf{98.4} \\

\midrule

\multicolumn{9}{l}{\textit{JEPA Models}} \\

V-JEPA2 ViT-L
& V & \textemdash & \textemdash & \textemdash & \textemdash & \underline{0.94} & \textemdash & \textemdash \\

V-JEPA2 ViT-g
& V & \textemdash & \textemdash & \textemdash & \textemdash & 1.02 & \textemdash & \textemdash \\

\midrule

\multicolumn{9}{l}{\textit{MLLMs}} \\

Qwen3.5-27B
& T, I, V & \textemdash & \textemdash & \textemdash & \textemdash & \textemdash & \textbf{61.3} & 92.0 \\

Qwen3.5-35B-A3B
& T, I, V & \textemdash & \textemdash & \textemdash & \textemdash & \textemdash & 58.8 & 91.2 \\

Qwen3.6-27B
& T, I, V & \textemdash & \textemdash & \textemdash & \textemdash & \textemdash & \underline{60.2} & 91.9 \\

Qwen3.6-35B-A3B
& T, I, V & \textemdash & \textemdash & \textemdash & \textemdash & \textemdash & 59.0 & 90.9 \\

\bottomrule
\end{tabular}%
}
\end{table*}

%% file: Tables/appendix_family_mapping_table.tex
\scriptsize
\renewcommand{\arraystretch}{1.10}
\setlength{\tabcolsep}{2.8pt}

\begin{longtable}{@{}
>{\raggedright\arraybackslash}p{0.190\textwidth}
>{\centering\arraybackslash}p{0.125\textwidth}
>{\raggedright\arraybackslash}p{0.175\textwidth}
>{\raggedright\arraybackslash}p{0.325\textwidth}
>{\centering\arraybackslash}p{0.100\textwidth}@{}}
\caption{Summary of simulation families, their physical formulations, computation methods, case parameters, and field outputs.}
\label{tab:family_physics_mapping}\\

\toprule
Family & Formulation & Computation & Case parameters & Field output \\
\midrule
\endfirsthead

\toprule
Family & Formulation & Computation & Case parameters & Field output \\
\midrule
\endhead

\midrule
\multicolumn{5}{r}{Continued on next page} \\
\endfoot

\bottomrule
\endlastfoot
\addlinespace[4pt]
\rule{0pt}{2.8ex}\textbf{Fluid Mechanics} & & & & \\
\cmidrule(lr){1-5}
\texttt{buoyancy} &
~\eqref{eq:fluid-ns}, \eqref{eq:fluid-boussinesq} &
OpenFOAM (FVM) &
cavity dimensions; temperature difference; reference temperature; kinematic viscosity; thermal diffusivity; gravity &
$\lVert\mathbf{U}\rVert$ \\
\texttt{cavity} &
~\eqref{eq:fluid-ns} &
OpenFOAM (FVM) &
cavity dimensions; lid velocity; kinematic viscosity; Reynolds number $Re$ &
$\lVert\mathbf{U}\rVert$ \\
\texttt{external\_\allowbreak{}aero} &
~\eqref{eq:fluid-ns} &
OpenFOAM (FVM) &
airfoil chord; thickness scale; free-stream velocity; kinematic viscosity; Reynolds number $Re$ &
$\lVert\mathbf{U}\rVert$ \\
\texttt{external\_\allowbreak{}wake} &
~\eqref{eq:fluid-ns} &
OpenFOAM (FVM) &
obstacle diameter or width; free-stream velocity; kinematic viscosity; Reynolds number $Re$ &
$\lVert\mathbf{U}\rVert$ \\
\texttt{internal\_\allowbreak{}step} &
~\eqref{eq:fluid-ns}, \eqref{eq:fluid-rheology} &
OpenFOAM (FVM) &
channel dimensions; inlet velocity; Reynolds number $Re$; reference viscosity $\nu_0$; viscosity bounds $\nu_{\min},\nu_{\max}$; consistency $K$; exponent $n$; yield stress $\tau_0$ &
$\lVert\mathbf{U}\rVert$ \\
\texttt{jet\_\allowbreak{}axi} &
~\eqref{eq:fluid-ns} &
OpenFOAM (FVM) &
jet radius; outer radius; jet velocity; kinematic viscosity; Reynolds number $Re$ &
$\lVert\mathbf{U}\rVert$ \\
\texttt{scalar\_\allowbreak{}transport} &
~\eqref{eq:fluid-ns}, \eqref{eq:fluid-scalar} &
OpenFOAM (FVM) &
domain dimensions; carrier velocity; scalar diffusivity; P\'eclet number $Pe$; injection position and duration &
$T$ (transported scalar) \\
\addlinespace[4pt]
\rule{0pt}{2.8ex}\textbf{Solid Mechanics} & & & & \\
\cmidrule(lr){1-5}
\texttt{uniaxial\allowbreak{}Tension} &
~\eqref{eq:solid-linear} &
DOLFINx (FEM) &
$L_x$; $L_y$; neck position; neck width; Young's modulus $E$; Poisson ratio $\nu$; load magnitude &
$\lVert\mathbf{u}\rVert$ \\
\texttt{uniaxial\allowbreak{}Compression} &
~\eqref{eq:solid-linear} &
DOLFINx (FEM) &
$L_x$; $L_y$; Young's modulus $E$; Poisson ratio $\nu$; load magnitude; loading direction &
$\lVert\mathbf{u}\rVert$ \\
\texttt{cantilever\allowbreak{}Localization} &
~\eqref{eq:solid-linear} &
DOLFINx (FEM) &
$L_x$; $L_y$; localization-band position; Young's modulus $E$; Poisson ratio $\nu$; load magnitude; tip force direction &
$\lVert\mathbf{u}\rVert$ \\
\texttt{plate\allowbreak{}With\allowbreak{}Hole} &
~\eqref{eq:solid-linear} &
DOLFINx (FEM) &
hole radius; hole center $x$; hole center $y$; $L_x$; $L_y$; Young's modulus $E$; Poisson ratio $\nu$; load magnitude &
$\lVert\mathbf{u}\rVert$ \\
\texttt{pure\allowbreak{}Shear} &
~\eqref{eq:solid-linear} &
DOLFINx (FEM) &
$L_x$; $L_y$; Young's modulus $E$; Poisson ratio $\nu$; load magnitude; shear ratio &
$\lVert\mathbf{u}\rVert$ \\
\texttt{three\allowbreak{}Point\allowbreak{}Bending} &
~\eqref{eq:solid-linear} &
DOLFINx (FEM) &
support span; loading width; $L_x$; $L_y$; Young's modulus $E$; Poisson ratio $\nu$; load magnitude; center-drop ratio &
$\lVert\mathbf{u}\rVert$ \\
\texttt{torsion\allowbreak{}Shear} &
~\eqref{eq:solid-linear} &
DOLFINx (FEM) &
twist-axis position; warping coefficient; $L_x$; $L_y$; Young's modulus $E$; Poisson ratio $\nu$; load magnitude; twist magnitude &
$\lVert\mathbf{u}\rVert$ \\
\texttt{biaxial\allowbreak{}Tension} &
~\eqref{eq:solid-linear} &
DOLFINx (FEM) &
$L_x$; $L_y$; biaxial loading ratio; Young's modulus $E$; Poisson ratio $\nu$; load magnitude &
$\lVert\mathbf{u}\rVert$ \\
\texttt{contact\allowbreak{}Patch\allowbreak{}Compression} &
~\eqref{eq:solid-linear} &
DOLFINx (FEM) &
patch center; patch half-width; $L_x$; $L_y$; Young's modulus $E$; Poisson ratio $\nu$; load magnitude; patch-load magnitude &
$\lVert\mathbf{u}\rVert$ \\
\texttt{cube\allowbreak{}Compression\allowbreak{}3D} &
~\eqref{eq:solid-linear} &
DOLFINx (FEM) &
$L_x$; $L_y$; $L_z$; Young's modulus $E$; Poisson ratio $\nu$; load magnitude; compression direction &
$\lVert\mathbf{u}\rVert$ \\
\texttt{long\allowbreak{}Cuboid\allowbreak{}Bending\allowbreak{}3D} &
~\eqref{eq:solid-linear} &
DOLFINx (FEM) &
$L_x$; $L_y$; $L_z$; Young's modulus $E$; Poisson ratio $\nu$; load magnitude &
$\lVert\mathbf{u}\rVert$ \\
\texttt{cylinder\allowbreak{}Torsion\allowbreak{}3D} &
~\eqref{eq:solid-linear} &
DOLFINx (FEM) &
cylinder radius $R$; $L_x$; $L_y$; $L_z$; Young's modulus $E$; Poisson ratio $\nu$; twist angle; load magnitude &
$\lVert\mathbf{u}\rVert$ \\
\texttt{sphere\allowbreak{}Indentation\allowbreak{}3D} &
~\eqref{eq:solid-linear} &
DOLFINx (FEM) &
sphere radius $R$; $L_x$; $L_y$; $L_z$; Young's modulus $E$; Poisson ratio $\nu$; indentation-depth ratio; load magnitude &
$\lVert\mathbf{u}\rVert$ \\
\texttt{plate\allowbreak{}With\allowbreak{}Hole\allowbreak{}3D} &
~\eqref{eq:solid-linear} &
DOLFINx (FEM) &
hole radius; hole center $x$; hole center $y$; hole center $z$; $L_x$; $L_y$; $L_z$; Young's modulus $E$ &
$\lVert\mathbf{u}\rVert$ \\
\texttt{notched\allowbreak{}Bar\allowbreak{}Tension\allowbreak{}3D} &
~\eqref{eq:solid-linear} &
DOLFINx (FEM) &
notch-depth ratio; notch-width ratio; notch position; $L_x$; $L_y$; $L_z$; Young's modulus $E$; Poisson ratio $\nu$ &
$\lVert\mathbf{u}\rVert$ \\
\texttt{neo\allowbreak{}Hookean} &
~\eqref{eq:solid-finite-strain} &
DOLFINx (nonlinear FEM) &
$L_x$; $L_y$; shear modulus $\mu$; volumetric coefficient $\kappa$; load magnitude &
$\lVert\mathbf{u}\rVert$ \\
\texttt{buckling\allowbreak{}Snapthrough} &
~\eqref{eq:solid-finite-strain} &
DOLFINx (nonlinear FEM) &
$L_x$; $L_y$; initial imperfection; Young's modulus $E$; Poisson ratio $\nu$; load magnitude; critical-load parameter &
$\lVert\mathbf{u}\rVert$ \\
\texttt{near\allowbreak{}Incompressible\allowbreak{}Block} &
~\eqref{eq:solid-finite-strain} &
DOLFINx (nonlinear FEM) &
$L_x$; $L_y$; shear modulus $\mu$; volumetric coefficient $\kappa$; near-incompressibility bias; load magnitude &
$\lVert\mathbf{u}\rVert$ \\
\texttt{hyperelastic\allowbreak{}Neo\allowbreak{}Hookean\allowbreak{}Cuboid\allowbreak{}3D} &
~\eqref{eq:solid-finite-strain} &
DOLFINx (nonlinear FEM) &
$L_x$; $L_y$; $L_z$; shear modulus $\mu$; volumetric coefficient $\kappa$; load magnitude &
$\lVert\mathbf{u}\rVert$ \\
\texttt{hyperelastic\allowbreak{}Neo\allowbreak{}Hookean\allowbreak{}3D} &
~\eqref{eq:solid-finite-strain} &
DOLFINx (nonlinear FEM) &
cylinder radius $R$; $L_x$; $L_y$; $L_z$; shear modulus $\mu$; volumetric coefficient $\kappa$; load magnitude &
$\lVert\mathbf{u}\rVert$ \\
\texttt{hyperelastic\allowbreak{}Neo\allowbreak{}Hookean\allowbreak{}Sphere\allowbreak{}3D} &
~\eqref{eq:solid-finite-strain} &
DOLFINx (nonlinear FEM) &
sphere radius $R$; $L_x$; $L_y$; $L_z$; shear modulus $\mu$; volumetric coefficient $\kappa$; load magnitude &
$\lVert\mathbf{u}\rVert$ \\
\texttt{yield\allowbreak{}Hardening} &
~\eqref{eq:solid-history} &
DOLFINx (nonlinear FEM) &
$L_x$; $L_y$; neck position; neck width; Young's modulus $E$; Poisson ratio $\nu$; yield stress $\sigma_y$; hardening modulus $H$ &
$\lVert\mathbf{u}\rVert$ \\
\texttt{load\allowbreak{}Unload\allowbreak{}Hysteresis} &
~\eqref{eq:solid-history} &
DOLFINx (nonlinear FEM) &
$L_x$; $L_y$; neck position; neck width; Young's modulus $E$; Poisson ratio $\nu$; yield stress $\sigma_y$; hardening modulus $H$ &
$\lVert\mathbf{u}\rVert$ \\
\texttt{necking} &
~\eqref{eq:solid-history} &
DOLFINx (nonlinear FEM) &
$L_x$; $L_y$; neck position; neck width; Young's modulus $E$; Poisson ratio $\nu$; yield stress $\sigma_y$; hardening modulus $H$ &
$\lVert\mathbf{u}\rVert$ \\
\texttt{plastic\allowbreak{}J\allowbreak{}2Cylinder\allowbreak{}Tension\allowbreak{}3D} &
~\eqref{eq:solid-history} &
DOLFINx (nonlinear FEM) &
cylinder radius $R$; $L_x$; $L_y$; $L_z$; Young's modulus $E$; Poisson ratio $\nu$; yield stress $\sigma_y$; hardening modulus $H$ &
$\lVert\mathbf{u}\rVert$ \\
\texttt{plastic\allowbreak{}J\allowbreak{}2Notched\allowbreak{}Bar\allowbreak{}3D} &
~\eqref{eq:solid-history} &
DOLFINx (nonlinear FEM) &
notch-depth ratio; notch-width ratio; notch position; $L_x$; $L_y$; $L_z$; Young's modulus $E$; Poisson ratio $\nu$ &
$\lVert\mathbf{u}\rVert$ \\
\texttt{plastic\allowbreak{}J\allowbreak{}2Large\allowbreak{}Strain\allowbreak{}3D} &
~\eqref{eq:solid-history} &
DOLFINx (nonlinear FEM) &
$L_x$; $L_y$; $L_z$; Young's modulus $E$; Poisson ratio $\nu$; yield stress $\sigma_y$; hardening modulus $H$; load magnitude &
$\lVert\mathbf{u}\rVert$ \\
\addlinespace[4pt]
\rule{0pt}{2.8ex}\textbf{Dynamics} & & & & \\
\cmidrule(lr){1-5}
\texttt{pendulum} &
~\eqref{eq:dyn-rigid} &
Chrono &
link length; bob radius; mass; initial angle; initial angular velocity; gravity $g$; damping &
$\lVert\mathbf{v}\rVert$ on bob \\
\texttt{rigid\allowbreak{}Collision} &
~\eqref{eq:dyn-impact} &
Analytical / ODE integration &
body shape; body dimensions; body offset; mass; restitution coefficient $e$; friction coefficient; initial velocity; impact offset &
$\lVert\mathbf{v}\rVert$ on bodies \\
\texttt{sphere\allowbreak{}Drop\allowbreak{}3D} &
~\eqref{eq:dyn-rigid} &
Chrono &
sphere radius $R$; drop height; mass; restitution coefficient $e$; friction coefficient; initial velocity; gravity $g$; ground contact &
$\lVert\mathbf{v}\rVert$ on sphere \\
\texttt{stack\allowbreak{}Stability} &
~\eqref{eq:dyn-rigid} &
Chrono &
block count; block dimensions; stack offset; mass; friction coefficient; restitution coefficient $e$; disturbance velocity; gravity $g$ &
$\lVert\mathbf{v}\rVert$ on blocks \\
\texttt{block\allowbreak{}Tower\allowbreak{}3D} &
~\eqref{eq:dyn-rigid} &
Chrono &
tower height; block dimensions; impactor dimensions; mass; friction coefficient; restitution coefficient $e$; impact velocity; gravity $g$ &
$\lVert\mathbf{v}\rVert$ on bodies \\
\texttt{drop\allowbreak{}Weight\allowbreak{}Impact\allowbreak{}SMC} &
~\eqref{eq:dyn-contact} &
Analytical / ODE integration &
plate size and thickness; impactor radius and mass; contact stiffness $k_n$; contact damping $c_n$; plate stiffness, damping, and mass; drop height; initial velocity; gravity $g$ &
plate displacement $\lVert\mathbf{u}\rVert$ \\
\texttt{flexible\allowbreak{}Crank\allowbreak{}Slider} &
~\eqref{eq:dyn-crank} &
Analytical / ODE integration &
crank radius; connecting-rod length; slider-axis direction; initial phase; angular speed; rod-deflection amplitude &
rod deflection $u(s,t)$ \\
\texttt{gear\allowbreak{}Pair\allowbreak{}3D} &
~\eqref{eq:dyn-gear} &
Analytical / ODE integration &
gear count; gear radii; tooth counts; shaft locations; mesh clearance; initial phase; motor-speed profile &
angular speed $\lvert\omega\rvert$ \\
\texttt{cantilever\allowbreak{}Beam\allowbreak{}Transient} &
~\eqref{eq:dyn-cantilever} &
Analytical / ODE integration &
beam length $L$; width $b$; height $h$; Young's modulus $E$; density $\rho$; tip load $F$; load-removal time &
beam displacement $u$ \\
\texttt{harmonic\allowbreak{}Response\allowbreak{}FRF} &
~\eqref{eq:dyn-frf} &
Analytical / ODE integration &
frequency sweep; reference frequency $f_1$; force amplitude $F_0$; damping coefficient; drive frequency; beam length $L$ &
displacement $u$ \\
\texttt{reissner\allowbreak{}Shell\allowbreak{}Modal} &
~\eqref{eq:dyn-plate} &
Analytical / ODE integration &
plate dimensions $a,b$; thickness $t$; Young's modulus $E$; Poisson ratio $\nu$; density $\rho$; mode indices $(m,n)$; mode amplitude &
modal displacement $w$ \\
\texttt{stress\allowbreak{}Wave\allowbreak{}Propagation} &
~\eqref{eq:dyn-wave} &
Analytical / ODE integration &
rod length $L$; diameter; Young's modulus $E$; density $\rho$; pulse amplitude; rise and fall times &
axial stress $\sigma_{xx}$ \\
\texttt{rubber\allowbreak{}Block\allowbreak{}Compression} &
~\eqref{eq:dyn-prescribed} &
Analytical / ODE integration &
block dimensions; mesh resolution; compression ratio; loading duration; effective stiffness &
displacement-derived scalar $u_{\mathrm{interp}}$ \\
\texttt{j\allowbreak{}2Plastic\allowbreak{}Beam} &
~\eqref{eq:dyn-prescribed} &
Analytical / ODE integration &
beam length $L$; element count; maximum tip deflection; residual tip deflection; maximum equivalent plastic strain &
equivalent plastic strain $\bar{\varepsilon}^{p}$ / displacement $u$ \\
\texttt{lemaitre\allowbreak{}Damage\allowbreak{}Bar} &
~\eqref{eq:dyn-prescribed} &
Analytical / ODE integration &
bar length $L$; node count; notch width; displacement amplitude; cycle frequency; damage-start cycle; failure cycle &
damage $D$ \\
\texttt{mass\allowbreak{}Spring\allowbreak{}Damper} &
~\eqref{eq:dyn-msd} &
Analytical / ODE integration &
number of degrees of freedom; mass spacing; cart count; mass; spring stiffness; damping; initial displacement; initial velocity &
displacement $x_i$ \\
\texttt{rolling\allowbreak{}Sliding\allowbreak{}Transition} &
~\eqref{eq:dyn-rolling} &
Analytical / ODE integration &
roller radius; roller width; track length; friction coefficient; rolling friction; inertia factor; initial translational speed; initial angular speed &
local surface speed $\lVert\mathbf{v}_s\rVert$ \\
\texttt{unbalanced\allowbreak{}Rotor\allowbreak{}3D} &
~\eqref{eq:dyn-rotor} &
Analytical / ODE integration &
shaft length; rotor radius; rotor mass; critical speed; spin-speed profile; damping ratio $\zeta$; eccentricity; axis height &
lateral displacement $u_{\perp}$ \\
\texttt{articulated\allowbreak{}Chain} &
~\eqref{eq:dyn-chain} &
Analytical / ODE integration &
link count; link lengths; initial joint angles; damping; base-drive amplitude and frequency; phase lag; terminal mass &
link / joint speed $\lVert\mathbf{v}\rVert$ \\
\addlinespace[4pt]
\rule{0pt}{2.8ex}\textbf{Optics \& EM} & & & & \\
\cmidrule(lr){1-5}
\texttt{plane\allowbreak{}Mirror\allowbreak{}Reflection} &
~\eqref{eq:opt-reflection} &
Analytical / ODE integration &
incidence angle; mirror reflectance; screen distance; beam divergence &
$I(x,y)$ (irradiance) \\
\texttt{concave\allowbreak{}Mirror\allowbreak{}Focus} &
~\eqref{eq:opt-imaging} &
Analytical / ODE integration &
focal length; aperture radius; object distance; detector distance &
$I(x,y)$ (irradiance) \\
\texttt{dielectric\allowbreak{}Block\allowbreak{}Refraction} &
~\eqref{eq:opt-refraction} &
Analytical / ODE integration &
incidence angle; refractive index $n$; block thickness; screen distance &
$I(x,y)$ (irradiance) \\
\texttt{total\allowbreak{}Internal\allowbreak{}Reflection} &
~\eqref{eq:opt-refraction} &
Analytical / ODE integration &
incidence angle; critical angle; core index $n_1$; cladding index $n_2$ &
$I(x,y)$ (irradiance) \\
\texttt{fresnel\allowbreak{}Brewster} &
~\eqref{eq:opt-fresnel} &
Analytical / ODE integration &
incidence angle; Brewster angle; refractive index $n_2$; polarization &
$I(x,y)$ (irradiance) \\
\texttt{beam\allowbreak{}Splitter\allowbreak{}Plate} &
~\eqref{eq:opt-fresnel} &
Analytical / ODE integration &
incidence angle; reflectance $R$; transmittance $T$; plate thickness &
$I(x,y)$ (irradiance) \\
\texttt{single\allowbreak{}Slit\allowbreak{}Fraunhofer} &
~\eqref{eq:opt-propagation} &
HCIPy &
slit width $a$; wavelength $\lambda$; screen distance; source offset &
$I(x,y)$ (irradiance) \\
\texttt{double\allowbreak{}Slit\allowbreak{}Interference} &
~\eqref{eq:opt-propagation} &
HCIPy &
slit width $a$; slit separation $d$; wavelength $\lambda$; relative phase $\phi$ &
$I(x,y)$ (irradiance) \\
\texttt{circular\allowbreak{}Aperture\allowbreak{}Airy} &
~\eqref{eq:opt-propagation} &
HCIPy &
aperture diameter $D$; wavelength $\lambda$; screen distance; aperture scale &
$I(x,y)$ (irradiance) \\
\texttt{knife\allowbreak{}Edge\allowbreak{}Fresnel} &
~\eqref{eq:opt-propagation} &
HCIPy &
edge position; wavelength $\lambda$; propagation distance $z$; beam width &
$I(x,y)$ (irradiance) \\
\texttt{diffraction\allowbreak{}Grating} &
~\eqref{eq:opt-interference} &
Analytical / ODE integration &
grating pitch $d$; slit count $N$; wavelength $\lambda$; screen distance &
$I(x,y)$ (irradiance) \\
\texttt{thin\allowbreak{}Film\allowbreak{}Interference} &
~\eqref{eq:opt-interference} &
Analytical / ODE integration &
film thickness $t$; film index $n$; incidence angle; wavelength band &
$I(x,y)$ (irradiance) \\
\texttt{gaussian\allowbreak{}Beam\allowbreak{}Propagation} &
~\eqref{eq:opt-gaussian} &
Analytical / ODE integration &
beam waist $w_0$; wavelength $\lambda$; propagation distance $z$; waist position &
$I(x,y)$ (irradiance) \\
\texttt{prism\allowbreak{}Dispersion} &
~\eqref{eq:opt-refraction} &
Analytical / ODE integration &
prism apex angle; wavelength-dependent refractive indices; incidence angle; prism length &
$I(x,y)$ (irradiance) \\
\texttt{thin\allowbreak{}Lens\allowbreak{}Focus} &
~\eqref{eq:opt-imaging} &
Analytical / ODE integration &
focal length; aperture radius; object distance; detector distance &
$I(x,y)$ (irradiance) \\
\texttt{glass\allowbreak{}Sphere\allowbreak{}Caustic} &
~\eqref{eq:opt-refraction} &
Analytical / ODE integration &
sphere radius $R$; refractive index $n$; detector distance; source distance &
$I(x,y)$ (irradiance) \\
\texttt{corner\allowbreak{}Cube\allowbreak{}Retroreflector} &
~\eqref{eq:opt-reflection} &
Analytical / ODE integration &
incidence angle; face-angle error; aperture radius; return-screen distance &
$I(x,y)$ (irradiance) \\
\texttt{polarizer\allowbreak{}Malus} &
~\eqref{eq:opt-polarization} &
Analytical / ODE integration &
analyzer angle; extinction ratio; input polarization; polarization mode &
$I(x,y)$ (irradiance) \\
\texttt{waveplate\allowbreak{}Polarization} &
~\eqref{eq:opt-polarization} &
Analytical / ODE integration &
retardance $\delta$; fast-axis angle; input polarization; analyzer angle &
$I(x,y)$ (irradiance) \\
\texttt{meep\allowbreak{}Cylinder\allowbreak{}Scattering} &
~\eqref{eq:em-maxwell} &
Meep (FDTD) &
cylinder radius $R$; refractive index $n$; source wavelength $\lambda$ &
normalized $Q(x,y,t)$ \\
\texttt{meep\allowbreak{}Waveguide\allowbreak{}Mode} &
~\eqref{eq:em-maxwell} &
Meep (FDTD) &
core width; core refractive index; source wavelength $\lambda$ &
normalized $Q(x,y,t)$ \\
\texttt{meep\allowbreak{}Photonic\allowbreak{}Crystal\allowbreak{}Defect} &
~\eqref{eq:em-maxwell} &
Meep (FDTD) &
lattice pitch; rod radius; defect scale; rod refractive index; source wavelength $\lambda$ &
normalized $Q(x,y,t)$ \\
\texttt{meep\allowbreak{}Photonic\allowbreak{}Crystal\allowbreak{}Waveguide} &
~\eqref{eq:em-maxwell} &
Meep (FDTD) &
lattice pitch; rod radius; rod refractive index; line-defect width; source wavelength $\lambda$ &
normalized $Q(x,y,t)$ \\
\texttt{meep\allowbreak{}Grating\allowbreak{}Diffraction} &
~\eqref{eq:em-maxwell} &
Meep (FDTD) &
grating period; duty cycle; bar refractive index; source wavelength $\lambda$ &
normalized $Q(x,y,t)$ \\
\texttt{meep\allowbreak{}Plasmonic\allowbreak{}Dimer} &
~\eqref{eq:em-maxwell}, \eqref{eq:em-drude} &
Meep (FDTD) &
particle radius; gap width; Drude frequency; Drude damping; source wavelength $\lambda$ &
normalized $Q(x,y,t)$ \\
\texttt{meep\allowbreak{}Dipole\allowbreak{}Radiation} &
~\eqref{eq:em-maxwell} &
Meep (FDTD) &
source wavelength $\lambda$; dipole orientation &
normalized $Q(x,y,t)$ \\
\texttt{meep\allowbreak{}Evanescent\allowbreak{}TIR} &
~\eqref{eq:em-maxwell} &
Meep (FDTD) &
incidence angle; high-index medium $n_1$; low-index medium $n_2$; source wavelength $\lambda$ &
normalized $Q(x,y,t)$ \\
\end{longtable}

%% file: Figures/query_example.tex
\begin{figure}[ht]
\centering

\caption{A template-generated retrieval query example \texttt{articulatedChain\_S300} from the training set. Colors indicate the functional components of the query, while \underline{underlining} marks case-specific values instantiated from the simulation record or derived diagnostics.}
\label{fig:query-highlight-example}

\vspace{5pt}

\begingroup
\setlength{\fboxsep}{4pt}
\setlength{\fboxrule}{0.35pt}

\fcolorbox{black}{white}{%
\begin{minipage}{0.94\linewidth}
\small

\textbf{Representative query from the released Dynamics data}
\hfill
{\footnotesize\texttt{articulatedChain\_S300}}

\vspace{5pt}

\begingroup
\setlength{\fboxsep}{3pt}
\setlength{\fboxrule}{0.4pt}

% \fcolorbox{lightgrayrule}{white}{%
\begin{minipage}{\dimexpr\linewidth-2\fboxsep-2\fboxrule\relax}
\justifying

\qscenario{Find the \slot{2D} dynamics video of \slot{3 link articulated pendulum chain}, showing a chain of hinged rigid links swinging under gravity or a root drive.}
\qmechanism{Physically, joint constraints transmit motion along the chain, creating phase lag, folding, or free-swing response.}
\qfield{The color map represents link and joint speed magnitude along the chain, so speed colors show which parts move fastest.}
\qconfig{Use the variant with drive mode set to \slot{free release}, link count set to \slot{short}, terminal mass set to \slot{low}; \slot{Earth-scale gravity} acts downward; \slot{no separate ground plane} is part of the visible setup.}
\qobs{To identify the exact video, look for the mechanism rotates through about \slot{25.03 degrees}, linked bodies reach a relative angle near \slot{25.55 degrees}, the fastest visible motion reaches about \slot{2.251 m/s}, and the recorded motion covers roughly \slot{6.57 s}.}

\end{minipage}%
% }
\endgroup

\vspace{7pt}

\centering
\footnotesize

\colorbox{qScenario}{\strut\hspace{7pt}}\, Scenario / visible response
\quad
\colorbox{qMechanism}{\strut\hspace{7pt}}\, Physical mechanism
\quad
\colorbox{qField}{\strut\hspace{7pt}}\, Visualized quantity

\vspace{3pt}

\colorbox{qConfig}{\strut\hspace{7pt}}\, Case-specific conditions
\quad
\colorbox{qObservable}{\strut\hspace{7pt}}\, Quantitative diagnostics

\end{minipage}%
}

\endgroup

\end{figure}

%% file: Tables/appendix_regression_targets.tex
\begin{table}[ht]
\caption{Continuous physical quantities used in the regression benchmark.}
\label{tab:regression_targets}
\centering
\small
\begin{tabular}{lll}
\toprule
Family & Target & Range \\
\midrule
\texttt{pendulum} & gravity $g$ & $2$--$18~\mathrm{m\,s^{-2}}$ \\
\texttt{rigidCollision} & mass $m_2$ & $2$--$18~\mathrm{kg}$ \\
\texttt{sphereDrop3D} & restitution $e$ & $0.35$--$0.90$ \\
\texttt{dye2d} & P\'eclet number $Pe$ & $30$--$1200$ \\
\texttt{flowpastcylinder2d} & Reynolds number $Re$ & $80$--$500$ \\
\texttt{totalInternalReflection} & refractive index $n$ & $1.44$--$1.78$ \\
\texttt{doubleSlitInterference} & wavelength $\lambda$ & $450$--$700~\mathrm{nm}$ \\
\texttt{cylinderTorsion3D} & shear modulus $G$ & $2$--$10~\mathrm{GPa}$ \\
\texttt{longCuboidBending3D} & Young's modulus $E$ & $5$--$50~\mathrm{GPa}$ \\
\texttt{hyperelasticNeoHookeanCuboid3D} & shear modulus $\mu$ & $5$--$25~\mathrm{MPa}$ \\
\bottomrule
\end{tabular}
\end{table}